\documentclass[graybox,envcountchap,sectrefs]{svmono}

\usepackage{mathptmx}
\usepackage{helvet}
\usepackage{courier}
\usepackage{type1cm}         
\usepackage{makeidx}         
\usepackage{graphicx}        
\usepackage{multicol}        
\usepackage[bottom]{footmisc}
\usepackage[english]{babel}
\usepackage{caption}
\usepackage{amsfonts}
\usepackage{psfrag}
\usepackage{verbatim}
\usepackage{amssymb}
\usepackage{graphicx,times,amsmath} 
\usepackage{array}
\usepackage{morefloats}
\usepackage{multirow}
\usepackage{url}
\usepackage{harvard} 
\usepackage[lined,ruled]{algorithm2e}
\usepackage{booktabs}

\makeindex             

\begin{document}

\author{Rocco Langone, Raghvendra Mall, Carlos Alzate, Johan A. K. Suykens}
\title{Kernel Spectral Clustering and applications}
\subtitle{Chapter Contribution to the book: Unsupervised Learning Algorithms}
\maketitle

\frontmatter

\include{dedic}
\include{foreword}
\include{preface}
\include{acknow}

\tableofcontents

\include{acronym}

\mainmatter
\include{part}
%
%
%
\chapter{Kernel Spectral Clustering and applications}
\label{ch:langonetall} 

\abstract{In this chapter we review the main literature related to kernel spectral clustering (KSC), an approach to clustering cast within a kernel-based optimization setting. KSC represents a least-squares support vector machine based formulation of spectral clustering described by a weighted kernel PCA objective. Just as in the classifier case, the binary clustering model is expressed by a hyperplane in a high dimensional space induced by a kernel. In addition, the multi-way clustering can be obtained by combining a set of binary decision functions via an Error Correcting Output Codes (ECOC) encoding scheme. Because of its model-based nature, the KSC method encompasses three main steps: training, validation, testing. In the validation stage model selection is performed to obtain tuning parameters, like the number of clusters present in the data. This is a major advantage compared to classical spectral clustering where the determination of the clustering parameters is unclear and relies on heuristics. Once a KSC model is trained on a small subset of the entire data, it is able to generalize well to unseen test points. Beyond the basic formulation, sparse KSC algorithms based on the Incomplete Cholesky Decomposition (ICD) and $L_0$, $L_1, L_0 + L_1$, Group Lasso regularization are reviewed. In that respect, we show how it is possible to handle large scale data. Also, two possible ways to perform hierarchical clustering and a soft clustering method are presented. Finally, real-world applications such as image segmentation, power load time-series clustering, document clustering and big data learning are considered. 
}

\section{Introduction}
\label{intro}
Spectral clustering (SC) represents the most popular class of algorithms based on graph theory \cite{chung}. It makes use of the Laplacian's spectrum to partition a graph into weakly connected sub-graphs. Moreover, if the graph is constructed based on any kind of data (vector, images etc.), data clustering can be performed\footnote{In this case the given data points represent the node of the graph and their similarity the corresponding edges.}. SC began to be popularized when Shi and Malik  introduced the Normalized Cut criterion to handle image segmentation \cite{malik}. Afterwards, Ng and Jordan \cite{ngjordan} in a theoretical work based on matrix perturbation theory have shown conditions under which a good performance of the algorithm is expected. Finally, in the tutorial by Von Luxburg the main literature related to SC has been exhaustively summarized \cite{luxburg_tutorial}. Although very successful in a number of applications, SC has some limitations. For instance, it cannot handle big data without using approximation methods like the Nystr\"om algorithm \cite{sc_nystrom,nystrom}, the power iteration method \cite{pic}, or linear algebra based methods \cite{Ning_PR,Dhanjal_arXiv,vanbarel_ICD}. Furthermore, the generalization to out-of-sample data is only approximate.

These issues have been recently tackled by means of a spectral clustering algorithm formulated as weighted kernel PCA \cite{multiway_pami}. The technique, named kernel spectral clustering (KSC), is based on solving a constrained optimization problem in a primal-dual setting. In other words, KSC is a Least Squares Support Vector Machine (LS-SVM \cite{lssvm_book}) model used for clustering instead of classification\footnote{This is a considerable novelty, since SVMs are typically known as classifiers or function approximation models rather than clustering techniques.}. By casting SC in a learning framework, KSC allows to rigorously select tuning parameters such as the natural number of clusters which are present in the data. Also, an accurate prediction of the cluster memberships for unseen points can be easily done by projecting test data in the embedding eigenspace learned during training. Furthermore, the algorithm can be tailored to a given application by using the most appropriate kernel function. Beyond that, by using sparse formulations and a fixed-size \cite{lssvm_book,fixed_size} approach, it is possible to readily handle big data. Finally, by means of adequate adaptations of the core algorithm, hierarchical clustering and a soft clustering approach have been proposed.  

All these topics will be detailed in the next Sections. Precisely, after presenting the basic KSC method, the soft KSC algorithm will be summarized. Next, two possible ways to accomplish hierarchical clustering will be explained. Afterwards, some sparse formulations based on the Incomplete Cholesky Decomposition (ICD) and $L_0$, $L_1, L_0 + L_1$, Group Lasso regularization will be described. Lastly, various interesting applications in different domains such as computer vision, power-load consumer profiling, information retrieval and big data clustering will be illustrated. All these examples assume a static setting. Concerning other applications in a dynamic scenario the interested reader can refer to \cite{rocco_singapore,journal_POM} for fault detection, to \cite{IKSC_paper} for incremental time-series clustering, to \cite{Langone_physicA,rocco_ICMSQUARE2012,MKSC_Orlando} in case of community detection in evolving networks and \cite{diego_IJCNN2013} in relation to human motion tracking.   

\section{Notation}
\label{notation}

\begin{center}
\begin{tabular}{p{.28\textwidth}p{.70\textwidth}}                                   
$x^T$ & Transpose of the vector $x$\\
$A^T$ & Transpose of the matrix $A$\\
$I_N$ & $N \times N$ Identity matrix\\
$1_N$ & $N \times 1$ Vector of ones\\
$\mathcal{D}_{\textrm{tr}} = \{x_i\}_{i=1}^{N_{\textrm{tr}}}$ & Training sample of $N_{\textrm{tr}}$ data points\\
$\varphi(\cdot)$ & Feature map\\
$\mathcal{F}$ & Feature space of dimension $d_h$\\
$\{\mathcal{A}_p\}_{p=1}^k$ & Partitioning composed of $k$ clusters\\
$\mathcal{G = (V,E)}$ & Set of $N$ vertices $\mathcal{V}= \{v_i\}_{i=1}^N$ and $m$ edges $\mathcal{E}$ of a graph\\
$\vert \cdot \vert$ & Cardinality of a set  
\end{tabular} 
\end{center} 

\section{Kernel Spectral Clustering (KSC)}
\label{sec:ksc}

\subsection{Mathematical formulation}
\label{KSC_model}

\subsubsection{Training problem}
The KSC formulation for $k$ clusters is stated as a combination of $k-1$ binary problems \cite{multiway_pami}. In particular, given a set of training data $\mathcal{D}_{\textrm{tr}} = \{x_i\}_{i=1}^{N_{\textrm{tr}}}$, the primal problem is:  
\begin{equation}
\label{primalKSC}
\begin{aligned}
& \underset{w^{(l)},e^{(l)},b_l}{\text{min}}
& & \frac{1}{2}\sum_{l=1}^{k-1}w^{(l)^T}w^{(l)}-\frac{1}{2}\sum_{l=1}^{k-1}\gamma_l e^{(l)^T}Ve^{(l)}\\
& \text{subject to}
& & e^{(l)}=\Phi w^{(l)}+b_l1_{N_{\textrm{tr}}}, l=1,\hdots,k-1.
\end{aligned}
\end{equation}
The $e^{(l)}=[e^{(l)}_1,\hdots, e^{(l)}_i,\hdots, e^{(l)}_{N_{\textrm{tr}}}]^T$ are the projections of the training data mapped in the feature space along the direction $w^{(l)}$. For a given point $x_i$, the model in the primal form is:
\begin{equation}
	\label{primalModel}
	e_i^{(l)} = w^{(l)^{T}}\varphi(x_i) +b_l.
\end{equation}
The primal problem (\ref{primalKSC}) expresses the maximization of the weighted variances of the data given by $e^{(l)^T}Ve^{(l)}$ and the contextual minimization of the squared norm of the vector $w^{(l)}$, $\forall l$. The regularization constants $\gamma_l\in\mathbb{R}^+$ mediate the model complexity expressed by $w^{(l)}$ with the correct representation of the training data. $V \in \mathbb{R}^{N_{\textrm{tr}} \times N_{\textrm{tr}}}$ is the weighting matrix and $\Phi$ is the $N_{\textrm{tr}}\times d_h$ feature matrix $\Phi=[\varphi(x_1)^T;\hdots;\varphi(x_{N_{\textrm{tr}}})^T]$, where $\varphi:\mathbb{R}^d\rightarrow\mathbb{R}^{d_h}$ denotes the mapping to a high-dimensional feature space, $b_l$ are bias terms.  

The dual problem corresponding to the primal formulation (\ref{primalKSC}), by setting $V = D^{-1}$ becomes\footnote{By choosing $V = I$, problem (\ref{dual_KSC}) is identical to kernel PCA \cite{suykens_KPCA,scholkopf_kpca1,mikakernelpca}.}:  
\begin{equation}
\label{dual_KSC}
D^{-1}M_D\Omega\alpha^{(l)}=\lambda{_l}\alpha^{(l)} 
\end{equation}
where $\Omega$ is the kernel matrix with $ij$-th entry $\Omega_{ij}=K(x_i,x_j) = \varphi(x_i)^T \varphi(x_j)$. $K:\mathbb{R}^d \times \mathbb{R}^d \rightarrow\mathbb{R}$ means the kernel function. The type of kernel function to utilize is application-dependent, as it is outlined in Table \ref{table_kernels}. The matrix $D$ is the graph degree matrix which is diagonal with positive elements $D_{ii}=\sum_j\Omega_{ij}$, $M_D$ is a centering matrix defined as $M_D = I_{N_{\textrm{tr}}} - \frac{1}{1_{N_{\textrm{tr}}}^TD^{-1}1_{N_{\textrm{tr}}}}1_{N_{\textrm{tr}}}1_{N_{\textrm{tr}}}^TD^{-1}$, the $\alpha^{(l)}$ are vectors of dual variables, $\lambda{_l} = \frac{N_{\textrm{tr}}}{\gamma_l}$, $K: \mathbb{R}^d \times \mathbb{R}^d \rightarrow \mathbb{R}$ is the kernel function. The dual clustering model for the $i$-th point can be expressed as follows:
 \begin{equation}
	\label{dualModel}
	e_i^{(l)}=\sum_{j=1}^{N_{\textrm{tr}}} \alpha_j^{(l)}K(x_j,x_i) +b_l, j = 1,\hdots,N_{\textrm{tr}}, l=1,\hdots,k-1.
\end{equation}
The cluster prototypes can be obtained by binarizing the projections $e_i^{(l)}$ as $\textrm{sign}(e_i^{(l)})$. This step is straightforward because, thanks to presence of the bias term $b_l$, both the $e^{(l)}$ and the $\alpha^{(l)}$ variables get automatically centred around zero. The set of the most frequent binary indicators form a code-book $\mathcal{CB} = \{c_p\}_{p = 1}^k$, where each code-word of length $k-1$ represents a cluster.

\begin{table}[htbp]
\begin{center}
\begin{tabular}{|c|c|c|} 
\hline 
\textbf{Application}  & \textbf{Kernel Name} &  \textbf{Mathematical Expression} \\
\hline
Vector data & RBF & $K(x_i,x_j) = \exp(-||x_i-x_j||_2^2/\sigma^2)$\\
\hline
Images & RBF$_{\chi^2}$ & $K(h^{(i)},h^{(j)}) = \exp(-\dfrac{\chi_{ij}^2}{\sigma_{\chi}^2})$\\
\hline
Text & Cosine  & $K(x_i,x_j) = \frac{x_i^Tx_j}{||x_i||||x_j||}$\\
\hline
Time-series & RBF$_{\textrm{cd}}$  & $K(x_i,x_j) = \textrm{exp}(-||x_i-x_j||^2_{\textrm{cd}}/\sigma_{\textrm{cd}}^2)$\\
\hline
\end{tabular}
\end{center}
\caption{\textbf{Types of kernel functions for different applications}. In this Table RBF means Radial Basis Function, $\sigma$ denotes the bandwidth of the kernel. The symbol $h^{(i)}$ indicates a color histogram representing the $i-$th pixel of an image, and to compare two histograms $h^{(i)}$ and $h^{(j)}$ the $\chi^2$ statistical test is used \protect\cite{chi2test}. Regarding time-series data, the symbol \textit{cd} means correlation distance 
\protect\cite{timeseries_clustering}, and $||x_i-x_j||_{\textrm{cd}} = \sqrt{\frac{1}{2}(1-R_{ij})}$, where $R_{ij}$ can indicate the Pearson or Spearman's rank correlation coefficient between time-series $x_i$ and $x_j$.}
\label{table_kernels}
\end{table}

Interestingly, problem (\ref{dual_KSC}) has a close connection with SC based on a random walk Laplacian. In this respect, the kernel matrix can be considered as a weighted graph $\mathcal{G = (V,E)}$ with the nodes $v_i \in \mathcal{V}$ represented by the data points $x_i$. This graph has a corresponding random walk in which the probability of leaving a vertex is distributed among the outgoing edges according to their weight: $p_{t+1} = Pp_t$, where  $P = D^{-1}\Omega$ indicates the transition matrix with the $ij$-th entry denoting the probability of moving from node $i$ to node $j$ in one time-step. Moreover, the stationary distribution of the Markov Chain describes the scenario where the random walker stays mostly in the same cluster and seldom moves to the other clusters \cite{meila_random,meila_random,meila_learning,JeanCharlesPNAS2010}.   

\subsubsection{Generalization}
\label{sc:OOSE}
Given the dual model parameters $\alpha^{(l)}$ and $b_l$, it is possible to assign a membership to unseen points by calculating their projections onto the eigenvectors computed in the training phase:
\begin{equation}
\label{OOSE}
e_{\textrm{test}}^{(l)} = \Omega_{\textrm{test}}\alpha^{(l)} + b_l1_{N_{\textrm{test}}} 
\end{equation}
where $\Omega_\textrm{test}$ is the $N_\textrm{test}\times N$ kernel matrix evaluated using the test points with entries $\Omega_\textrm{test,ri} = K(x_r^\textrm{test},x_i)$, $r = 1,\hdots,N_\textrm{test}$, $i = 1,\hdots,N_{\textrm{tr}}$. The cluster indicator for a given test point can be obtained by using an Error Correcting Output Codes (ECOC) decoding procedure:
\begin{itemize}
\item the score variable is binarized 
\item the indicator is compared with the training code-book $\mathcal{CB}$ (see previous Section), and the point is assigned to the nearest prototype in terms of Hamming distance.
\end{itemize} 

The KSC method, comprising training and test stage, is summarized in algorithm \ref{alg:KSC}, and the related Matlab package is freely available on the Web\footnote{\textit{http://www.esat.kuleuven.be/stadius/ADB/alzate/softwareKSClab.php}}.  

\begin{algorithm}[htbp]
\small
\LinesNumbered
\KwData{Training set $\mathcal{D}_{\textrm{tr}} = \{x_i\}_{i=1}^{N_{\textrm{tr}}}$, test set $\mathcal{D}_{\textrm{test}} = \{x_m^{\textrm{test}}\}_{m=1}^{N_{\textrm{test}}}$ kernel function $K: \mathbb{R}^d \times \mathbb{R}^d \rightarrow \mathbb{R}$ positive definite and localized ($K(x_i,x_j) \rightarrow 0$ if $x_i$ and $x_j$ belong to different clusters), kernel parameters (if any), number of clusters $k$.} 
\KwResult{Clusters $\{\mathcal{A}_1,\hdots,\mathcal{A}_k\}$, codebook $\mathcal{CB} = \{c_p\}_{p = 1}^k$ with $\{c_p\} \in \{-1,1\}^{k-1}$.}
compute the training eigenvectors $\alpha^{(l)}$, $l = 1,\hdots,k-1$, corresponding to the $k-1$ largest eigenvalues of problem (\ref{dual_KSC})\\
let $A \in \mathbb{R}^{N_{\textrm{tr}} \times (k-1)}$ be the matrix containing the vectors $\alpha^{(1)},\ldots,\alpha^{(k-1)}$ as columns\\
binarize $A$ and let the code-book $\mathcal{CB} = \{c_p\}_{p = 1}^k$ be composed by the $k$ encodings of $Q = \textrm{sign}(A)$ with the most occurrences\\ 
$\forall i$, $i=1,\hdots,N_{\textrm{tr}}$, assign $x_i$ to $A_{p^*}$ where $p^* = \textrm{argmin}_pd_H(\textrm{sign}(\alpha_i),c_p)$ and $d_H(.,.)$ is the Hamming distance\\
binarize the test data projections $\textrm{sign}(e_m^{(l)})$, $m = 1,\hdots,N_{\textrm{test}}$, and let $\textrm{sign}(e_m) \in \{-1,1\}^{k-1}$ be the encoding vector of $x_m^{\textrm{test}}$\\ 
$\forall m$, assign $x_m^{\textrm{test}}$ to $A_{p^*}$, where $p^* = \textrm{argmin}_pd_H(\textrm{sign}(e_m),c_p)$.  
\caption{KSC algorithm \protect\cite{multiway_pami} }
\label{alg:KSC}
\end{algorithm}

\subsubsection{Model selection}
\label{modsel_KSC}
\begin{figure}[htbp]
\centering
\begin{tabular}{cc}
\includegraphics[width=2.15in]{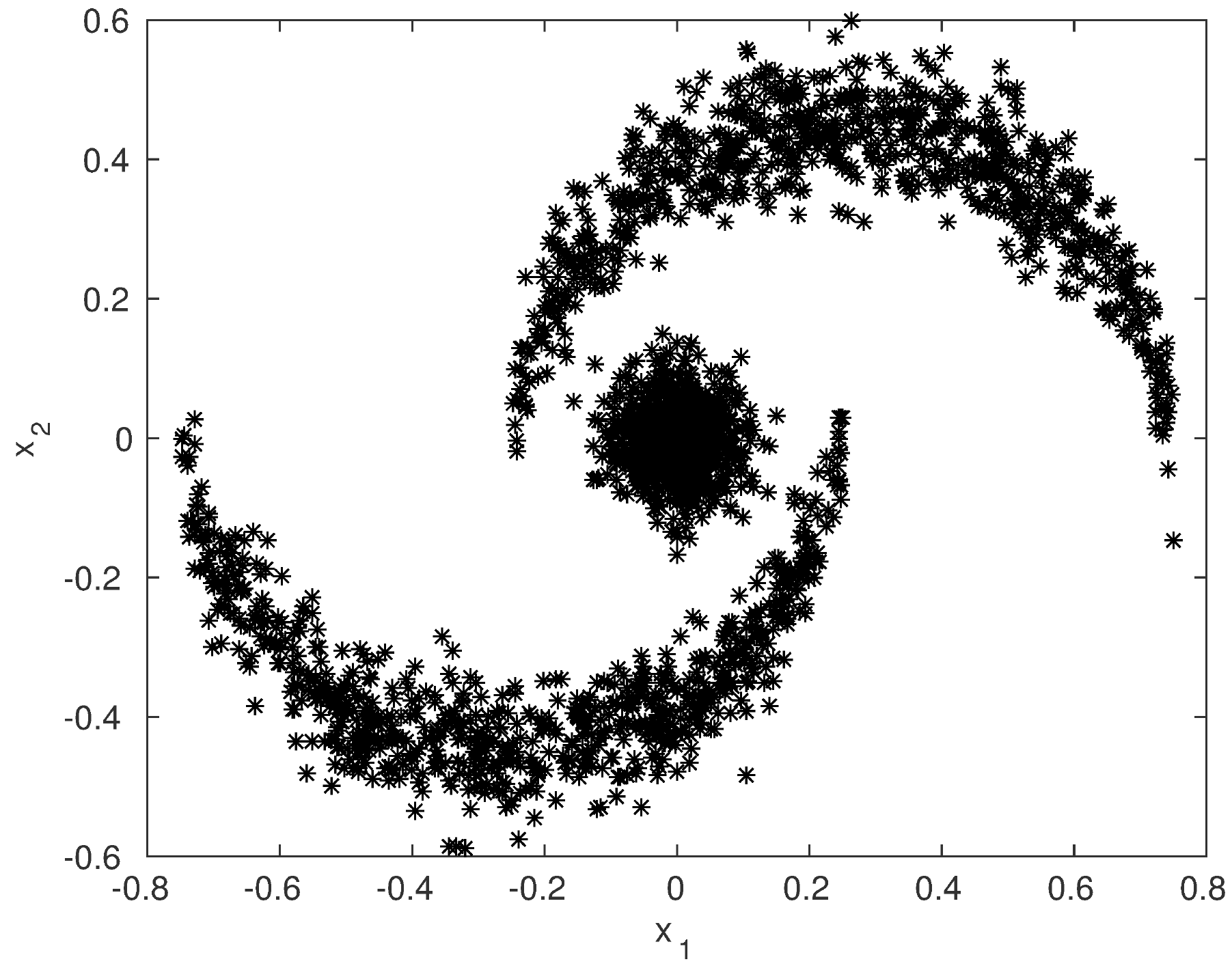}&\includegraphics[width=2.15in]{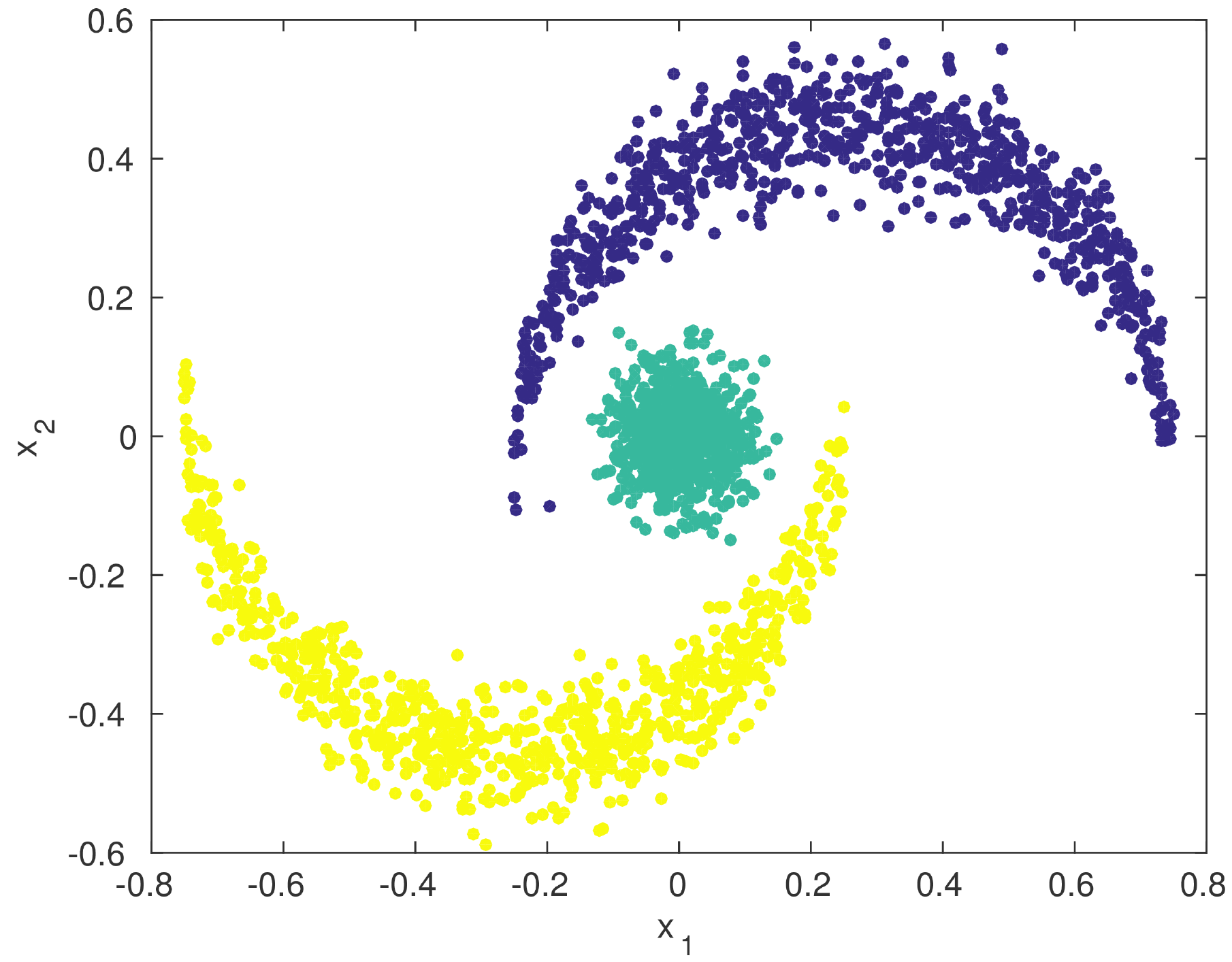}\\
\includegraphics[width=2.15in]{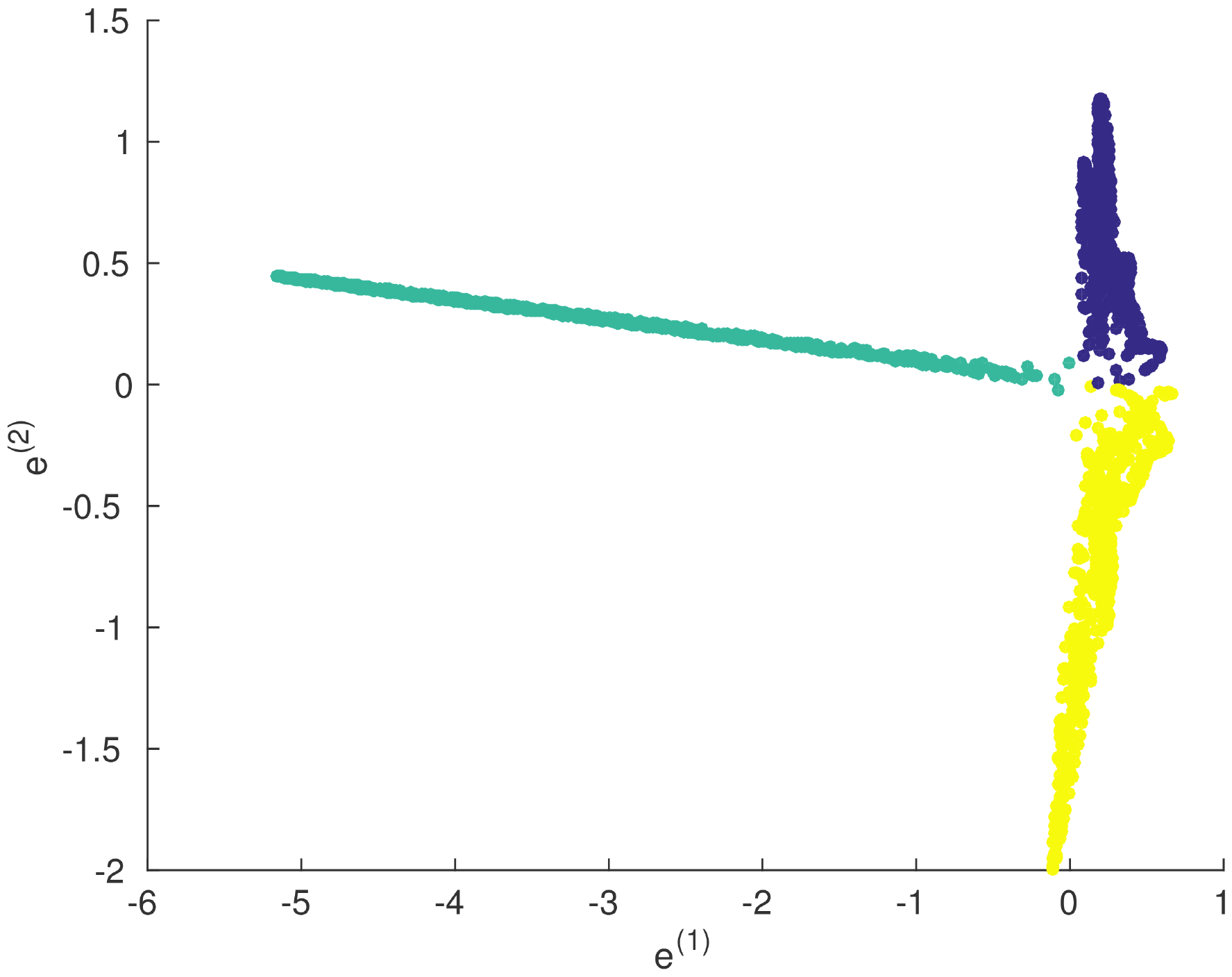}&\includegraphics[width=2.15in]{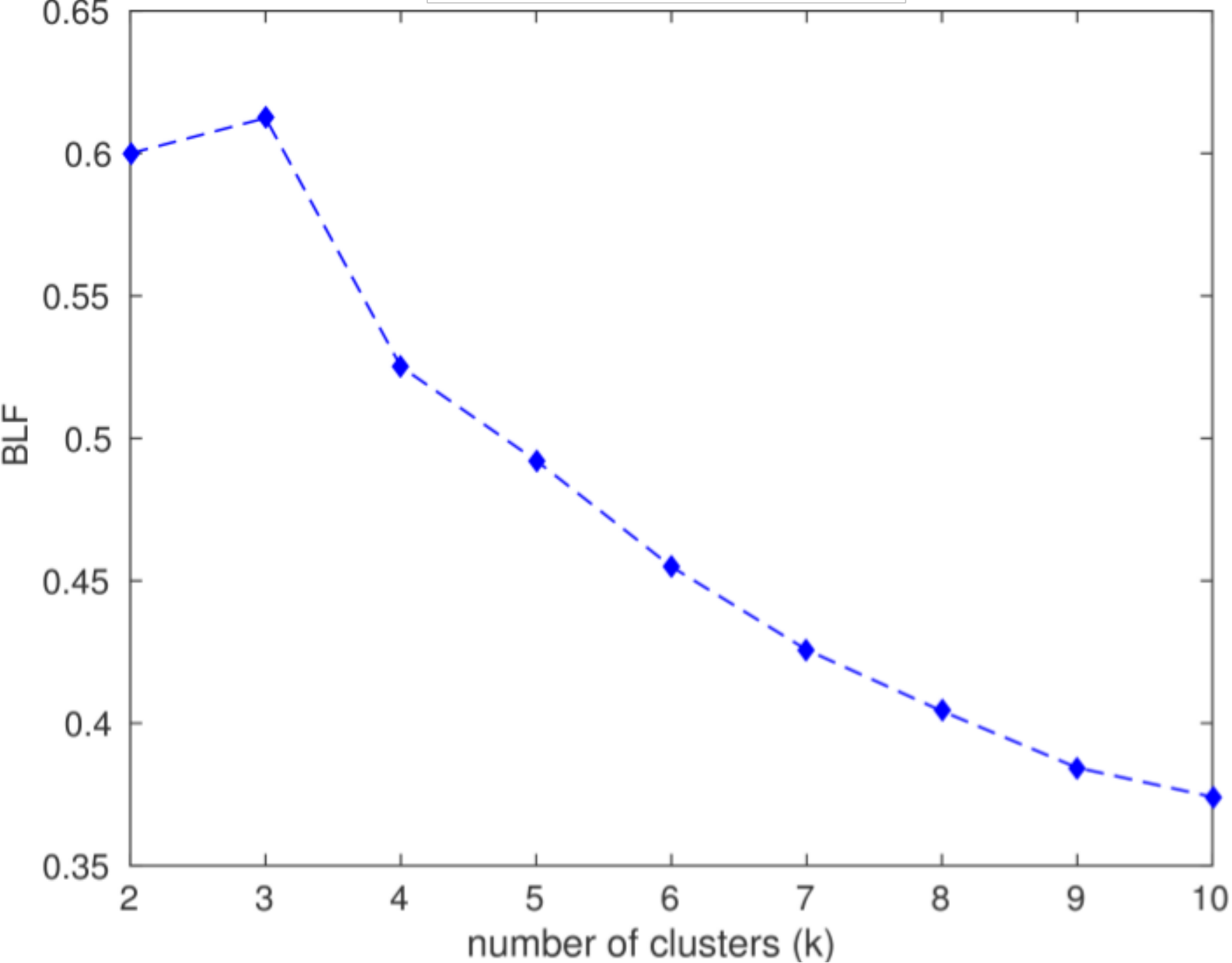}
\end{tabular}
\caption{\textbf{KSC partitioning on a toy dataset}. \textbf{(Top)} Original dataset consisting of $3$ clusters (left) and obtained clustering results (right). \textbf{(Bottom)} Points represented in the space of the projections $[e^{(1)},e^{(2)}]$ (left), for an optimal choice of $k$ (and $\sigma^2 = 4.36\cdot10^{-3}$) suggested by the BLF criterion (right). We can notice how the points belonging to one cluster tend to lie on the same line. A perfect line structure is not attained due to a certain amount of overlap between the clusters.}\label{clu_toy1}
\end{figure}

In order to select tuning parameters like the number of clusters $k$ and eventually the kernel parameters, a model selection procedure based on grid search is adopted. First, a validation set $\mathcal{D}_{\textrm{val}} = \{x_i\}_{i=1}^{N_{\textrm{val}}}$ is sampled from the whole dataset. Then, a grid of possible values of the tuning parameters is constructed. Afterwards, a KSC model is trained for each combination of parameters and the chosen criterion is evaluated on the partitioning predicted for the validation data. Finally, the parameters yielding  the maximum value of the criterion are selected. Depending on the kind of data, a variety of model selection criteria have been proposed:   
\begin{itemize}
\item \emph{Balanced Line Fit (BLF)}. It indicates the amount of collinearity between validation points belonging to the same cluster, in the space of the projections. It reaches its maximum value $1$ in case of well separated clusters, represented as lines in the space of the $e_{\textrm{val}}^{(l)}$ (see for instance the bottom left side of Figure \ref{clu_toy1})  
\item \emph{Balanced Angular Fit or BAF} \cite{raghvendra_bigdata}. For every cluster, the sum of the cosine similarity between the validation points and the cluster prototype, divided by the cardinality of that cluster, is computed. These similarity values are then summed up and divided by the total number of clusters.
\item \emph{Average Membership Strength abbr. AMS} \cite{rocco_IJCNN2013}. The mean membership per cluster denoting the mean degree of belonging of the validation points to the cluster is computed. These mean cluster memberships are then averaged over the number of clusters. 
\item \emph{Modularity} \cite{newmanmod2}. This quality function is well suited for network data. In the model selection scheme, the Modularity of the validation sub-graph corresponding to a given partitioning is computed, and the parameters related to the highest Modularity are selected \cite{rocco_IJCNN2011,rocco_IJCNN2012}.
\item \emph{Fisher Criterion}. The classical Fisher criterion \cite{Bishop2006} used in classification has been adapted to
select the number of clusters $k$ and the kernel parameters in the KSC framework \cite{AlzateHKS}. The criterion maximizes the distance between the means of the two clusters while minimizing the variance within each cluster, in the space of the projections $e_{\textrm{val}}^{(l)}$.   
\end{itemize}

In Figure \ref{clu_toy1} an example of clustering obtained by KSC on a synthetic dataset is shown. The BLF model selection criterion has been used to tune the bandwidth of the RBF kernel and the number of clusters. It can be noticed how the results are quite accurate, despite the fact that the clustering boundaries are highly nonlinear. 

\subsection{Soft Kernel Spectral Clustering (SKSC)}
\label{SKSC}
Soft kernel spectral clustering (SKSC) makes use of algorithm \ref{alg:KSC} in order to compute a first hard partitioning of the training data. Next, soft cluster assignments are performed by computing the cosine distance between each point and some cluster prototypes in the space of the projections $e^{(l)}$. In particular, given the projections for the training points $e_i = [e_i^{(1)},\hdots,e_i^{(k-1)}]$, $i=1,\hdots,N_{\textrm{tr}}$ and the corresponding hard assignments $q_i^p$ we can calculate for each cluster the cluster prototypes $s_1,\hdots,s_p,\hdots,s_k$, $s_p \in \mathbb{R}^{k-1}$ as:
\begin{equation}
\centering
\label{e_prototypes}
s_p = \frac{1}{n_p}\sum_{i=1}^{n_p}e_i 
\end{equation}             
where $n_p$ is the number of points assigned to cluster $p$ during the initialization step by KSC.
Then the cosine distance between the $i$-th point in the projections space and a prototype $s_p$ is calculated by means of the following formula:
\begin{equation}
\centering
\label{cosine_dist}
d^{\textrm{cos}}_{ip} = 1-e_i^Ts_p/(||e_i||_2||s_p||_2).
\end{equation}
The soft membership of point $i$ to cluster $q$ can be finally expressed as:
\begin{equation}
\centering
\label{soft_membership}
sm_i^{(q)} = \frac{\prod_{j\neq q}d^{\textrm{cos}}_{ij}}{\sum_{p=1}^k\prod_{j\neq p}d^{\textrm{cos}}_{ij}}
\end{equation}
with $\sum_{p=1}^{k}sm_i^{(p)} = 1$.             
As pointed-out in \cite{prob_clust_israel}, this membership represents a subjective probability expressing the belief in the clustering assignment.

The out-of-sample extension on unseen data consists simply of calculating eq. (\ref{OOSE}) and assigning the test projections to the closest centroid.

\begin{figure}[htbp]
\centering
\begin{tabular}{cc}
\includegraphics[width=2.85in]{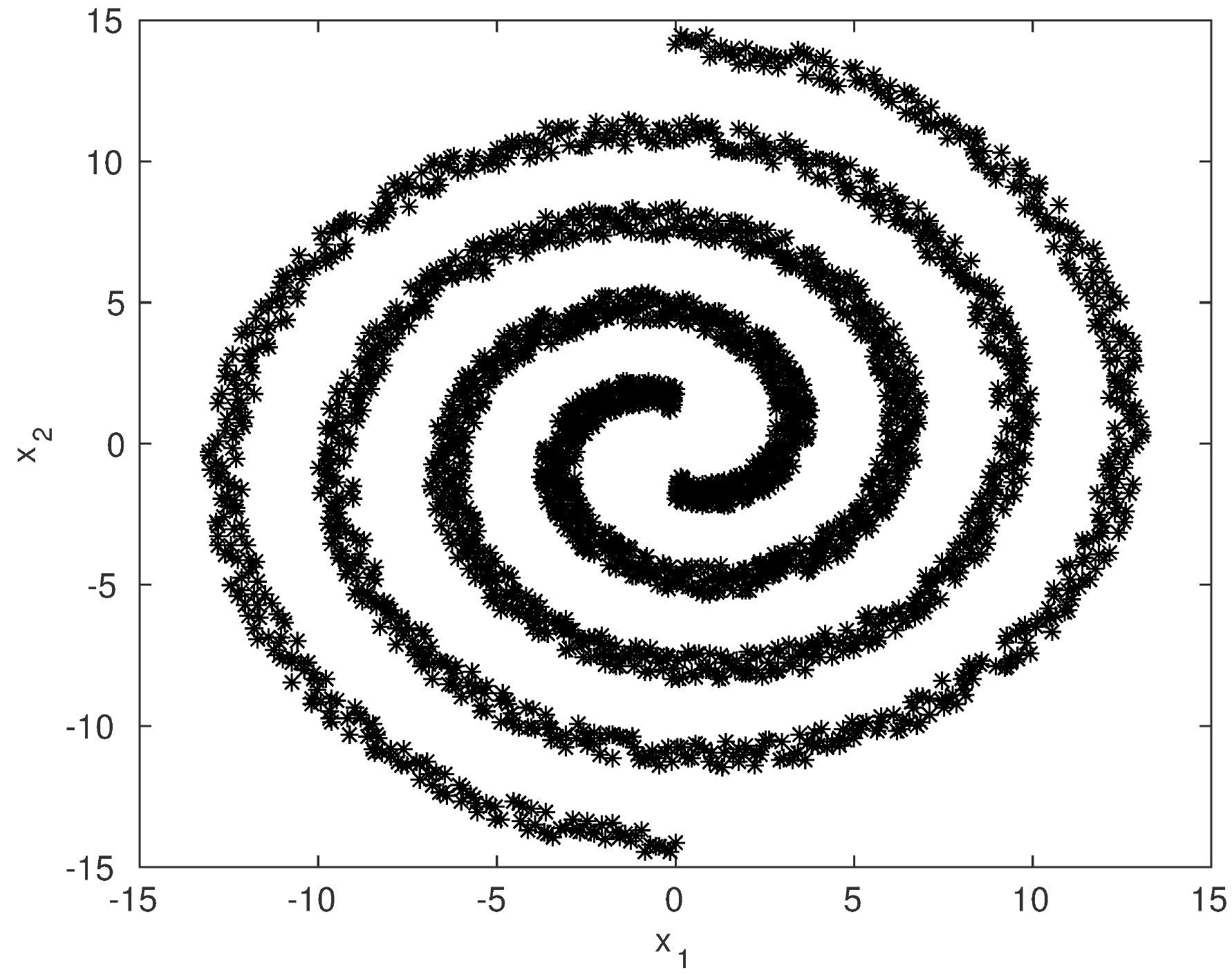}&\includegraphics[width=1.5in]{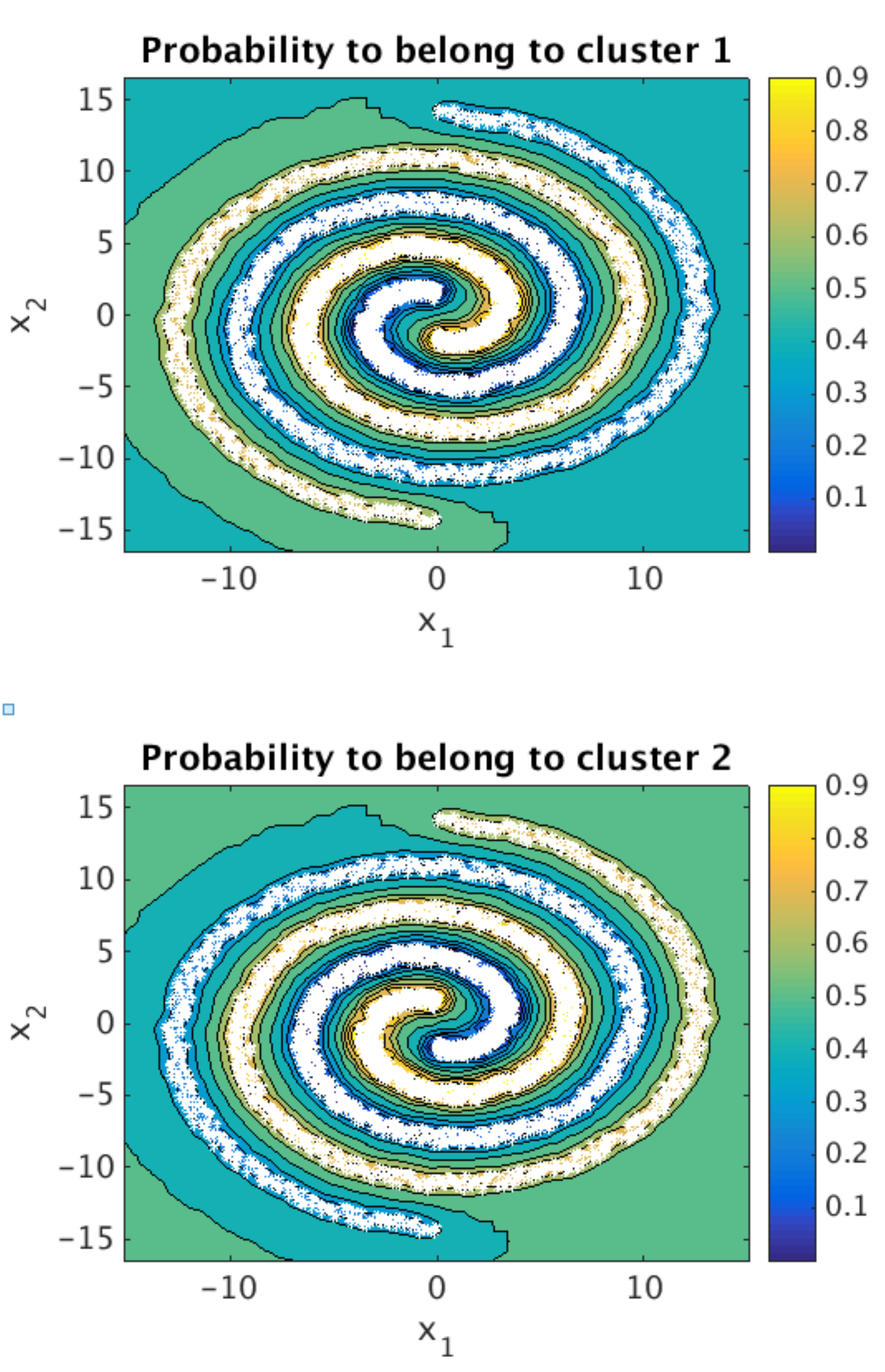}\\
\includegraphics[height=1.75in]{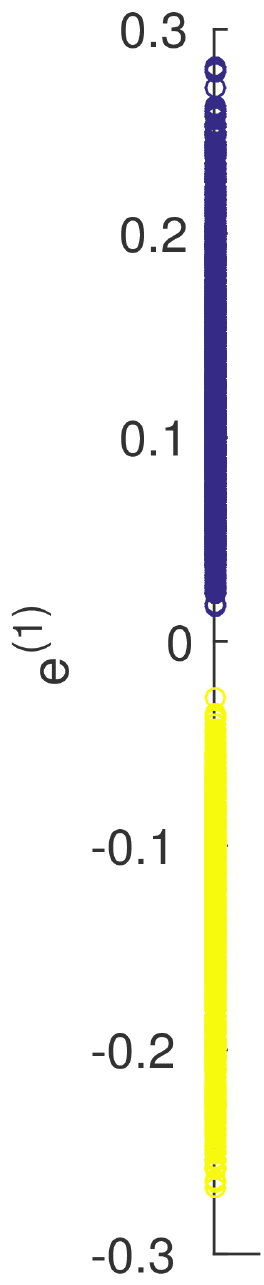}&\hspace{-2cm}\includegraphics[width=2.15in]{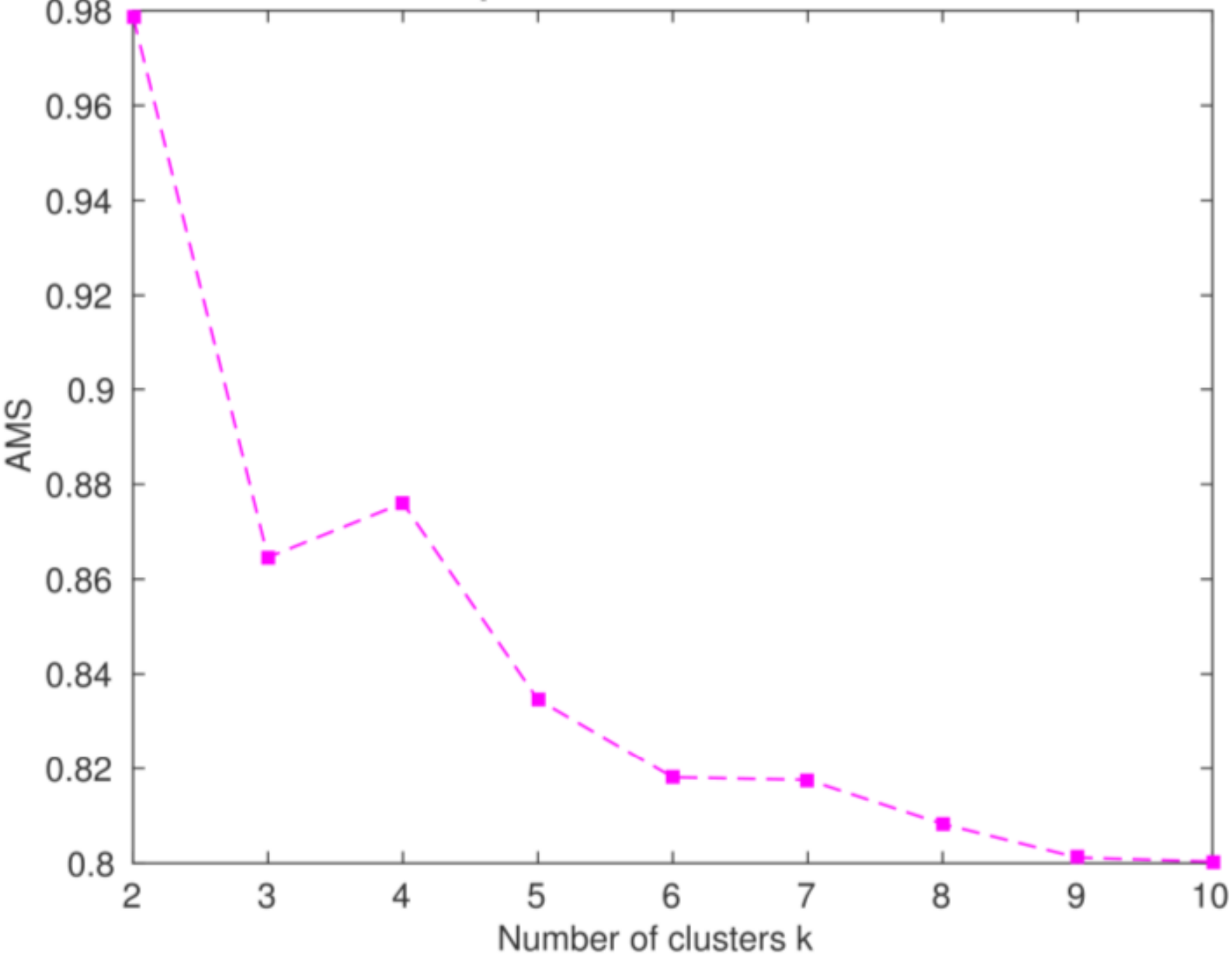}
\end{tabular}
\caption{\textbf{SKSC partitioning on a synthetic dataset}. \textbf{(Top)} Original dataset consisting of $2$ clusters (left) and obtained soft clustering results (right). \textbf{(Bottom)} Points represented in the space of the projection $e^{(1)}$ (left), for an optimal choice of $k$ (and $\sigma^2 = 1.53\cdot10^{-3}$) as detected by the AMS criterion (right). }\label{softclu_toy2}
\end{figure}

An example of soft clustering performed by SKSC on a synthetic dataset is depicted in Figure \ref{softclu_toy2}. The AMS  model selection criterion has been used to select the bandwidth of the RBF kernel and the optimal number of clusters. The reader can appreciate how SKSC provides more interpretable outcomes compared to KSC.

The SKSC method is summarized in algorithm \ref{alg:SKSC} and a Matlab implementation is freely downloadable\footnote{ \textit{http://www.esat.kuleuven.be/stadius/ADB/langone/softwareSKSClab.php}}.

\begin{algorithm}[htbp]
\small
\LinesNumbered
\KwData{Training set $\mathcal{D}_{\textrm{tr}} = \{x_i\}_{i=1}^{N_{\textrm{tr}}}$ and test set $\mathcal{D}_{\textrm{test}} = \{x_m^{\textrm{test}}\}_{m=1}^{N_{\textrm{test}}}$, kernel function $K: \mathbb{R}^d \times \mathbb{R}^d \rightarrow \mathbb{R}$ positive definite and localized ($K(x_i,x_j) \rightarrow 0$ if $x_i$ and $x_j$ belong to different clusters), kernel parameters (if any), number of clusters $k$.}
\KwResult{Clusters $\{\mathcal{A}_1,\hdots,\mathcal{A}_p,\hdots,\mathcal{A}_k\}$, soft cluster memberships $sm^{(p)}, p =1,\hdots,k$, cluster prototypes $\mathcal{SP} = \{s_p\}_{p = 1}^k$, $s_p \in \mathbb{R}^{k-1}$.} 
Initialization by solving eq. (\ref{dualModel}).\\ 
Compute the new prototypes $s_1,\hdots,s_k$ (eq. (\ref{e_prototypes})).\\ 
Calculate the test data projections $e_m^{(l)}$, $m = 1,\hdots,N_{\textrm{test}}$, $l = 1,\hdots,k-1$.\\
Find the cosine distance between each projection and all the prototypes (eq. (\ref{cosine_dist}))  
$\forall m$, assign $x_m^{\textrm{test}}$ to cluster $A_{p}$ with membership $sm^{(p)}$ according to eq. (\ref{soft_membership}).  
\caption{SKSC algorithm \protect\cite{rocco_IJCNN2013}}
\label{alg:SKSC}
\end{algorithm}

\subsection{Hierarchical Clustering}
\label{HC}
In many cases, clusters are formed by sub-clusters which in turn might have substructures. As a consequence, an algorithm able to discover a hierarchical organization of the clusters provides a more informative result, incorporating several scales in the analysis. The flat KSC algorithm has been extended in two ways in order to deal with hierarchical clustering. 

\subsubsection{Approach 1}
\label{HC1}

This approach, named hierarchical kernel spectral clustering (HKSC), was proposed in \cite{AlzateHKS} and exploits the information of a multi-scale structure present in the data given by the Fisher criterion (see end of Section \ref{modsel_KSC}). A grid search over different values of $k$ and $\sigma^2$ is performed to find tuning parameter pairs such that the criterion is greater than a specified threshold value. The KSC model is then trained for each pair and evaluated at the test set using the out-of-sample extension. A specialized linkage criterion determines which clusters are merging based on the evolution of the cluster memberships as the hierarchy goes up. The whole procedure is summarized in algorithm \ref{alg:HKSC}.

\begin{algorithm}[htbp]
\small
\LinesNumbered
\KwData{Training set $\mathcal{D}_{\textrm{tr}} = \{x_i\}_{i=1}^{N_{\textrm{tr}}}$, Validation set $\mathcal{D}_{\textrm{val}} = \{x_i\}_{i=1}^{N_{\textrm{val}}}$ and test set $\mathcal{D}_{\textrm{test}} = \{x_m^{\textrm{test}}\}_{m=1}^{N_{\textrm{test}}}$, RBF kernel function with parameter $ \sigma^2$, maximum number of clusters $k_{\textrm{max}}$, set of $R$ $\sigma^2$ values $\{\sigma^2_1,\hdots,\sigma^2_R\}$, Fisher threshold $\theta$.}
\KwResult{Linkage matrix $Z$} 
For every combination of parameter pairs $(k,\sigma^2)$ train a KSC model using algorithm \ref{alg:KSC}, predict the cluster memberships for validation points and calculate the related Fisher criterion\\ 
$ \forall k$, find the maximum value of the Fisher criterion across the given range of $\sigma^2$ values. If the maximum value is greater than the Fisher threshold $\theta$, create a set of these optimal $(k_*, \sigma^2_*)$ pairs.\\
Using the previously found $(k_*, \sigma^2_*)$ pairs train a clustering model and compute the cluster memberships for the test set using the out-of-sample extension.\\
Create the linkage matrix $Z$ by identifying which clusters merge starting from the bottom of the tree which contains max $k_*$ clusters.
\caption{HKSC algorithm \protect\cite{AlzateHKS}}
\label{alg:HKSC}
\end{algorithm}

\subsubsection{Approach 2}
\label{HC2}
\begin{figure}[htbp]
	\centering
	\includegraphics[width=\textwidth]{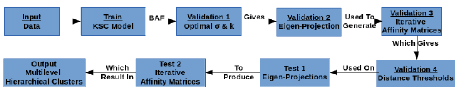}
	\caption{\textbf{AH-KSC algorithm}. Steps of AH-KSC method as described in \protect\cite{raghvendra_plosone} with addition of the step where the optimal $\sigma$ and $k$ are estimated.}
	\label{AHKSC2_fig}
\end{figure}

\begin{figure}[htbp]
	\centering
	\includegraphics[width=\textwidth, height = 2in]{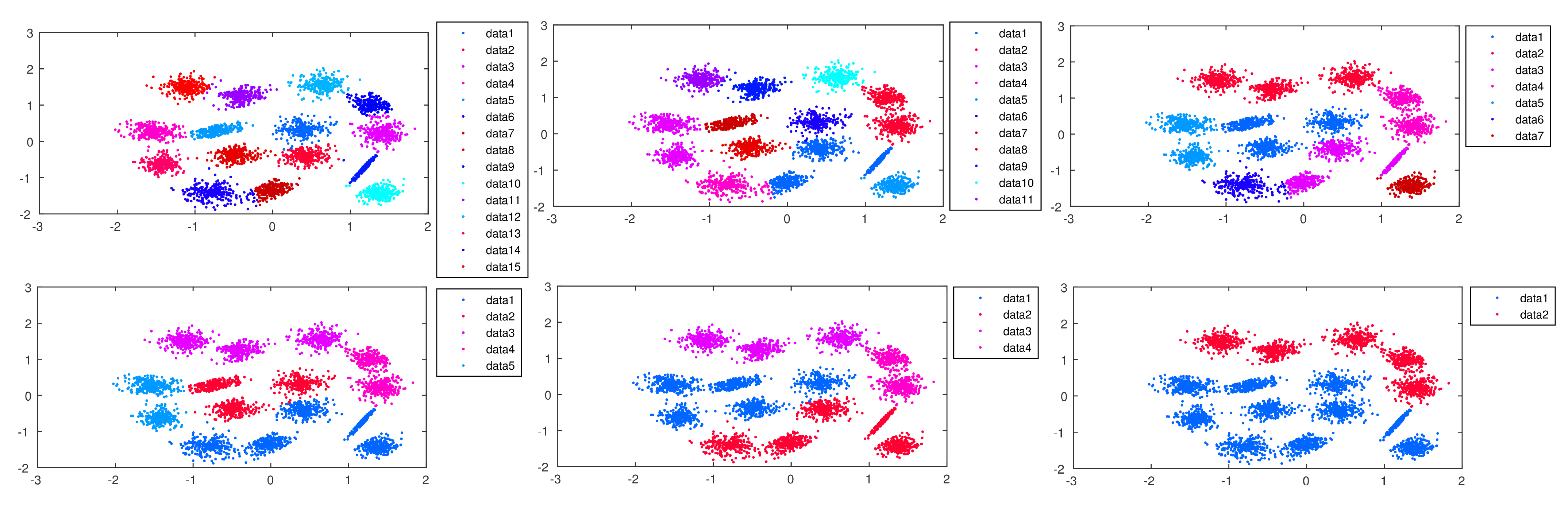}
	\caption{\textbf{AH-KSC partitioning on a toy dataset.} Cluster memberships for a toy dataset at different hierarchical levels obtained by the AH-KSC method.}\label{hclu_toy4}
\end{figure}

In \cite{raghvendra_plosone} and \cite{raghvendra_CIDM2014} an alternative hierarchical extension of the basic KSC algorithm was introduced, for network and vector data respectively. In this method, called agglomerative hierarchical kernel spectral clustering (AH-KSC), the structure of the projections in the eigenspace is used to automatically determine a set of increasing distance thresholds. At the beginning, the validation point with maximum number of similar points within the first threshold value is selected. The indices of all these points represent the first cluster at level $0$ of hierarchy. These points are then removed from the validation data matrix, and the process is repeated iteratively until the matrix becomes empty. Thus, the first level of hierarchy corresponding to the first distance threshold is obtained. To obtain the clusters at the next level of hierarchy the clusters at the previous levels are treated as data points, and the whole procedure is repeated again with other threshold values. This step takes inspiration from \cite{louvainmethod}. The algorithm stops when only one cluster remains. The same procedure is applied in the test stage, where the distance thresholds computed in the validation phase are used. An overview of all the steps involved in the algorithm is depicted in Figure \ref{AHKSC2_fig}. In Figure \ref{hclu_toy4} an example of hierarchical clustering performed by this algorithm on a toy dataset is shown.

\subsection{Sparse Clustering Models}
\label{sparse_KSC}
The computational complexity of the KSC algorithm depends on solving the eigenvalue problem (\ref{dual_KSC}) related to the training stage and computing eq. (\ref{OOSE}) which gives the cluster memberships of the remaining points. Assuming that we have $N_{\textrm{tot}}$ data and we use $N_{\textrm{tr}}$ points for training and $N_{\textrm{test}} = N_{\textrm{tot}} - N_{\textrm{tr}}$ as test set, the runtime of algorithm \ref{alg:KSC} is $O(N_{\textrm{tr}}^2) + O(N_{\textrm{tr}}N_{\textrm{test}})$. In order to reduce the computational complexity, it is then necessary to find a reduced set of training points, without loosing accuracy. In the next Sections two different methods to obtain a sparse KSC model, based on the Incomplete Cholesky Decomposition (ICD) and $L_1$ and $L_0$ penalties respectively, are discussed. In particular, thanks to the ICD, the KSC computational complexity for the training problem is decreased to $O(R^2N_{\textrm{tr}})$ \cite{Mihaly_ICD}, where $R$ indicates the reduced set size.     

\subsubsection{Incomplete Cholesky Decomposition}
\label{ICD_KSC}
One of the KKT optimality conditions characterizing the Lagrangian of problem (\ref{primalKSC}) is: 
\begin{equation}
\label{w_primal}
w^{(l)} = \Phi^T \alpha^{(l)} = \sum_{i=1}^{N_{\textrm{tr}}}\alpha_i^{(l)}\varphi(x_i).
\end{equation}
From eq. (\ref{w_primal}) it is evident that each training data point contributes to the primal variable $w^{(l)}$, resulting in a non-sparse model. In order to obtain a parsimonious model a reduced set method based on the Incomplete Cholesky Decomposition (ICD) was proposed in \cite{Carlos_ICD,Mihaly_ICD}. The technique is based on finding a small number $R \ll N_{\textrm{tr}}$ of points $\mathcal{R} = \{\hat{x_r}\}_{r=1}^{R}$ and related coefficients $\zeta^{(l)}$ with the aim of approximating $w^{(l)}$ as:
\begin{equation}
\label{w_sparse}
w^{(l)} \approx  \hat{w}^{(l)} = \sum_{r=1}^{R} \zeta_r^{(l)} \varphi(\hat{x_r}).
\end{equation}
As a consequence, the projection of an arbitrary data point $x$ into the training embedding is given by:
\begin{equation}
\label{e_sparse}
e^{(l)} \approx  \hat{e}^{(l)} = \sum_{r=1}^{R} \zeta_r^{(l)}K(x,\hat{x_r}) + \hat{b_l}.
\end{equation} 
The set $\mathcal{R}$ of points can be obtained by considering the pivots of the ICD performed on the kernel matrix $\Omega$. In particular, by assuming that $\Omega$ has a small numerical rank, the kernel matrix can be approximated by $\Omega \approx \hat{\Omega} = GG^T$, with $G \in \mathbb{R}^{N_{\textrm{tr}} \times R}$. If we plug in this approximated kernel matrix in problem (\ref{dual_KSC}), the KSC eigenvalue problem can be written as:
\begin{equation}
\label{dual_sparse}
\hat{D}^{-1} M_{\hat{D}} U \Psi ^2 U^T \hat{\alpha}^{(l)} = \hat{\lambda _l} \hat{\alpha}^{(l)}, l=1,\hdots,k
\end{equation}    
where $U \in \mathbb{R}^{N_{\textrm{tr}} \times R}$ and $V \in \mathbb{R}^{N_{\textrm{tr}} \times R}$ denotes the left and right singular vectors deriving from the singular value decomposition (SVD) of $G$, and $\Psi \in \mathbb{R}^{N_{\textrm{tr}} \times N_{\textrm{tr}}}$ is the matrix of the singular values. If now we pre-multiply both sides of eq. (\ref{dual_sparse}) by $U^T$ and replace $\hat{\delta}^{(l)} = U^T \hat{\alpha}^{(l)}$, only the following eigenvalue problem of size $ R \times R$ must be solved:
\begin{equation}
\label{dual_sparse2}
U^T\hat{D}^{-1}M_{\hat{D}}U \Psi^2 \hat{\delta}^{(l)} = \hat{\lambda _l}\hat{\delta}^{(l)}, l=1,\hdots,k.
\end{equation}  
The approximated eigenvectors of the original problem (\ref{dual_KSC}) can be computed as $ \hat{\alpha}^{(l)} = U \hat{\delta}^{(l)}$, and the sparse parameter vector can be found by solving the following optimization problem:
\begin{equation}
\label{sparse_ksc_eq}
min_{\zeta^{(l)}} \parallel w^{(l)} - \hat{w}^{(l)} \parallel^2 _2 = min_{\zeta^{(l)}} \parallel \Phi^T\alpha^{(l)} - \chi^T\zeta^{(l)} \parallel^2 _2.
\end{equation}
The corresponding dual problem can be written as follows:
\begin{equation}\label{recons_sksc}
\Omega^{\chi \chi}\delta^{(l)} = \Omega^{\chi\phi}\alpha^{(l)},
\end{equation}
where $\Omega^{\chi\chi}_{rs} = K(\tilde{x}_{r},\tilde{x}_{s})$, $\Omega^{\chi\phi}_{ri} = K(\tilde{x}_{r},x_{i})$, $r,s =1,\ldots,R, i=1,\ldots, N_{tr}$ and $l = 1,\ldots, k-1$.
Since the size $R$ of problem (\ref{dual_sparse2}) can be much smaller than the size $N_{\textrm{tr}}$ of the starting problem, the sparse KSC method\footnote{A \textit{C} implementation of the algorithm can be downloaded at:\\ \textit{http://www.esat.kuleuven.be/stadius/ADB/novak/softwareKSCICD.php}} is suitable for big data analytics.  

\subsubsection{Using Additional Penalty terms}
\label{L1_KSC}
In this part we explore sparsity in the KSC technique by using an additional penalty term in the objective function (\ref{sparse_ksc_eq}). In \cite{Carlos_ICD}, the authors used an $L_{1}$ penalization term in combination with the reconstruction error term to introduce sparsity. It is well known that the $L_{1}$ regularization introduces sparsity as shown in \cite{zhu:hastie}. However, the resulting reduced set is neither the sparsest nor the most optimal w.r.t. the quality of clustering for the entire dataset. In \cite{raghvendra_IJCNN2014}, we introduced alternative penalization techniques like Group Lasso \cite{yuan:lin} and \cite{friedman}, $L_{0}$ and $L_{1}+L_{0}$ penalizations. The Group Lasso penalty is ideal for clusters as it results in groups of relevant data points. The $L_{0}$ regularization calculates the number of non-zero terms in the vector. The $L_{0}$-norm results in a non-convex and NP-hard optimization problem. We modify the convex relaxation of $L_{0}$-norm based on an iterative re-weighted $L_{1}$ formulation introduced in \cite{boyd,hua:zheng}. We apply it to obtain the optimal reduced sets for sparse kernel spectral clustering. Below we provide the formulation for Group Lasso penalized objective (\ref{grouplasso_sksc}) and re-weighted $L_{1}$-norm penalized objectives (\ref{reweighted_sksc}).\\

The Group Lasso \cite{yuan:lin} based formulation for our optimization problem is:
\begin{equation}\label{grouplasso_sksc}
\begin{aligned}
& \underset{\beta \in \mathbb{R}^{N_{tr}\times (k-1)}}{\text{min}}
& & \| \Phi^\intercal \alpha -  \Phi^\intercal \beta \|_{2}^{2} + \lambda \sum_{l=1}^{N_{tr}} \sqrt{\rho_{l}}\|\beta_{l} \|_{2},
\end{aligned}
\end{equation}
where $\Phi = [\phi(x_{1}),\ldots,\phi(x_{N_{tr}})]$, $\alpha = [\alpha^{(1)},\ldots, \alpha^{(k-1)}]$, $\alpha \in \mathbb{R}^{N_{tr}\times (k-1)}$ and $\beta = [\beta_{1},\ldots,\beta_{N_{tr}}]$, $\beta \in \mathbb{R}^{N_{tr}\times (k-1)}$ . Here $\alpha^{(i)} \in \mathbb{R}^{N_{tr}}$ while $\beta_{j} \in \mathbb{R}^{k-1}$ and we set $\sqrt{\rho_{l}}$ as the fraction of training points belonging to the cluster to which the $l^{th}$ training point belongs. By varying the value of $\lambda$ we control the amount of sparsity introduced in the model as it acts as a regularization parameter. In \cite{friedman}, the authors show that if the initial solutions are $\hat{\beta}_{1}, \hat{\beta}_{2},\ldots, \hat{\beta}_{N_{tr}}$ then if $ \| X_{l}^\intercal( y - \sum_{i\neq l} X_{i}\hat{\beta}_{i}) \| < \lambda$, then $\hat{\beta}_{l}$ is zero otherwise it satisfies: $\hat{\beta}_{l} = (X_{l}^\intercal X_{l} + \lambda/\| \hat{\beta}_{l} \|)^{-1}X_{l}^\intercal r_{l}$ where $r_{l} = y - \sum_{i\neq l}X_{i} \hat{\beta}_{i}$.

Analogous to this, the solution to the group lasso penalization for our problem can be defined as: $\| \phi(x_{l})(\Phi^\intercal \alpha - \sum_{i \neq l} \phi(x_{i})\hat{\beta}_{i}) \| < \lambda$ then $\hat{\beta}_{l}$ is zero otherwise it satisfies: $\hat{\beta}_{l} = (\Phi ^\intercal \Phi + \lambda/\|\hat{\beta}_{l}\|)^{-1}\phi(x_{l})r_{l}$ where $r_{l} = \Phi^\intercal \alpha - \sum_{i \neq l} \phi(x_{i})\hat{\beta}_{i}$. The Group Lasso penalization technique can be solved by a blockwise co-ordinate  descent procedure as shown in \cite{yuan:lin}. The time complexity of the approach is $O(\text{maxiter}*k^{2}N_{tr}^2)$ where $\text{maxiter}$ is the maximum number of iterations specified for the co-ordinate descent procedure and $k$ is the number of clusters obtained via KSC. From our experiments we observed that on an average $10$ iterations suffice for convergence.

Concerning the re-weighted $L_{1}$ procedure, we modify the algorithm related to classification as shown in \cite{hua:zheng} and use it for obtaining the reduced set in our clustering setting:
\begin{equation}\label{reweighted_sksc}
\begin{aligned}
& \underset{\beta \in \mathbb{R}^{N_{tr}\times (k-1)}}{\text{min}}
& & \| \Phi^\intercal \alpha -  \Phi^\intercal \beta \|_{2}^{2} + \rho \sum_{i=1}^{N_{tr}} \epsilon_{i} + \| \Lambda\beta \|_{2}^{2} \\
& \text{such that}
& & \| \beta_{i} \|_{2}^{2} \leq \epsilon_{i}, i=1,\ldots, N_{tr}\\
& & & \epsilon_{i} \geq 0,
\end{aligned}
\end{equation}
where $\Lambda$ is matrix of the same size as the $\beta$ matrix i.e. $\Lambda \in \mathbb{R}^{N_{tr}\times(k-1)}$. The term $\| \Lambda\beta \|_{2}^{2}$ along with the constraint $\| \beta_{i} \|_{2}^{2} \leq \epsilon_{i}$ corresponds to the $L_{0}$-norm penalty on $\beta$ matrix. $\Lambda$ matrix is initially defined as a matrix of ones so that it gives equal chance to each element of $\beta$ matrix to reduce to zero. The constraints on the optimization problem forces each element of $\beta_{i} \in \mathbb{R}^{(k-1)}$ to reduce to zero. This helps to overcome the problem of sparsity per component which is explained in \cite{Carlos_ICD}. The $\rho$ variable is a regularizer which controls the amount of sparsity that is introduced by solving this optimization problem. 

\begin{figure}[htbp]
\centering
\begin{tabular}{c}
\includegraphics[width=2.5in,height=1.5in]{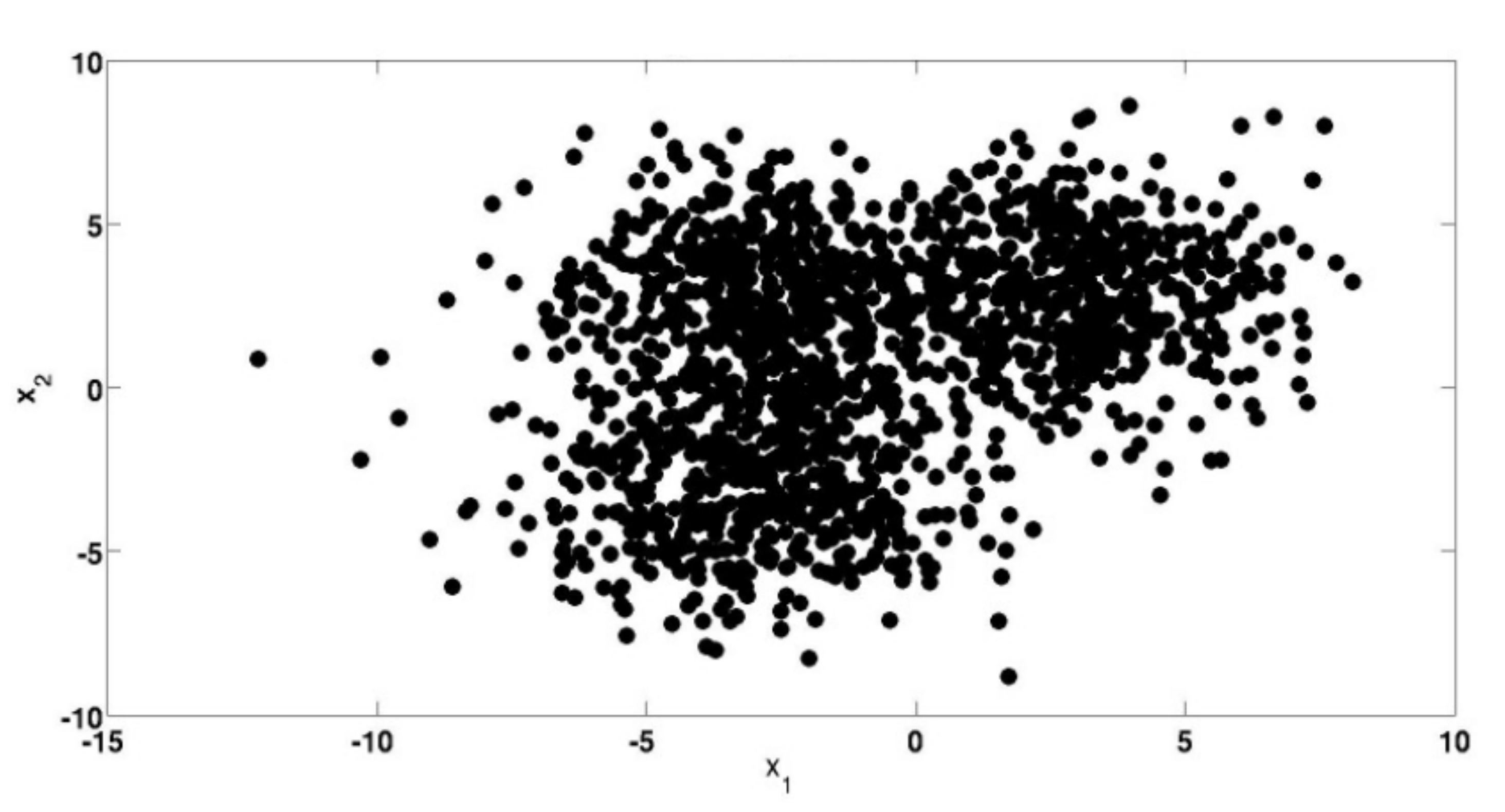}\\
\includegraphics[width=2.5in,height=1.5in]{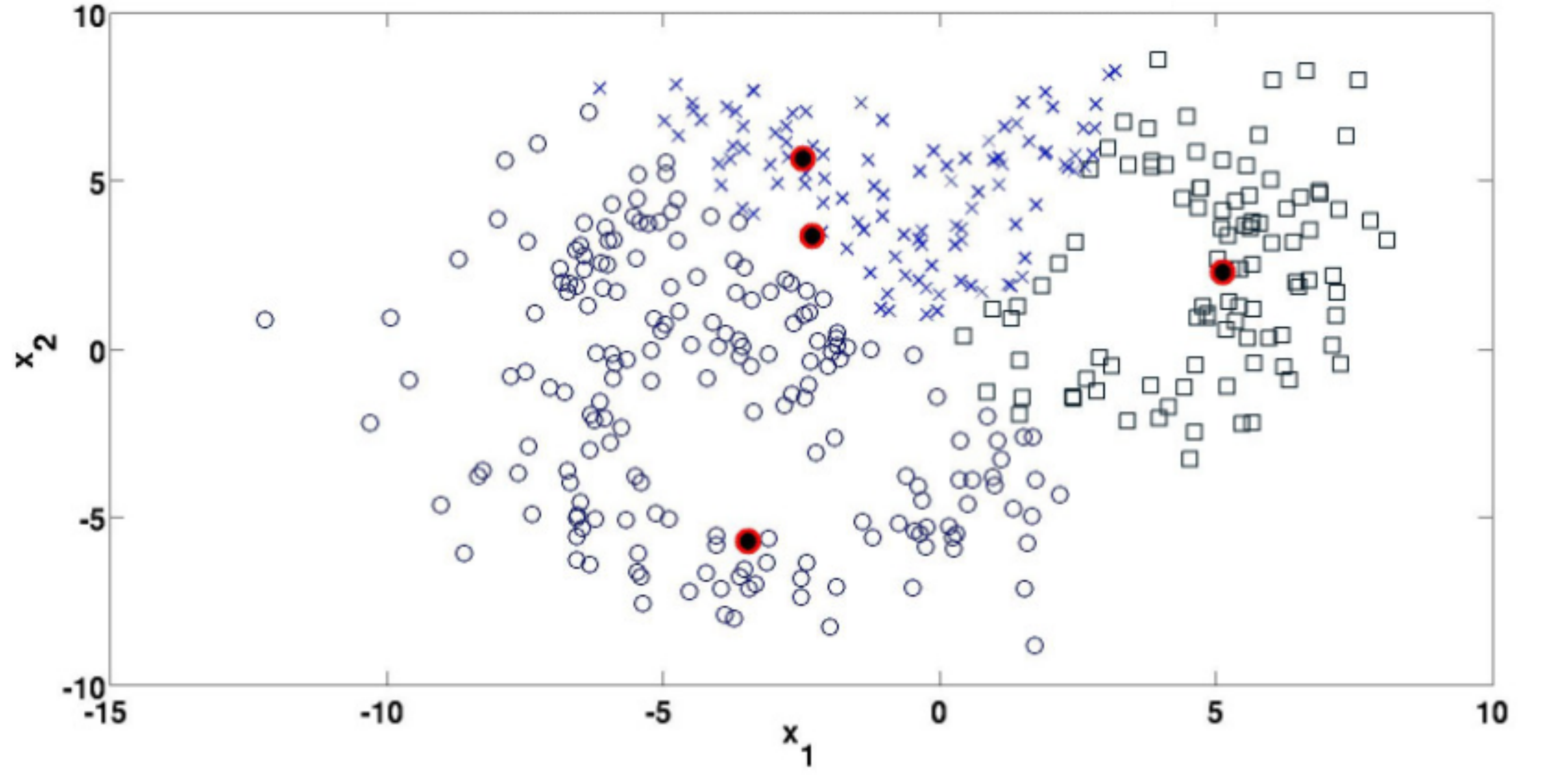}\\
\includegraphics[width=2.5in,height=1.5in]{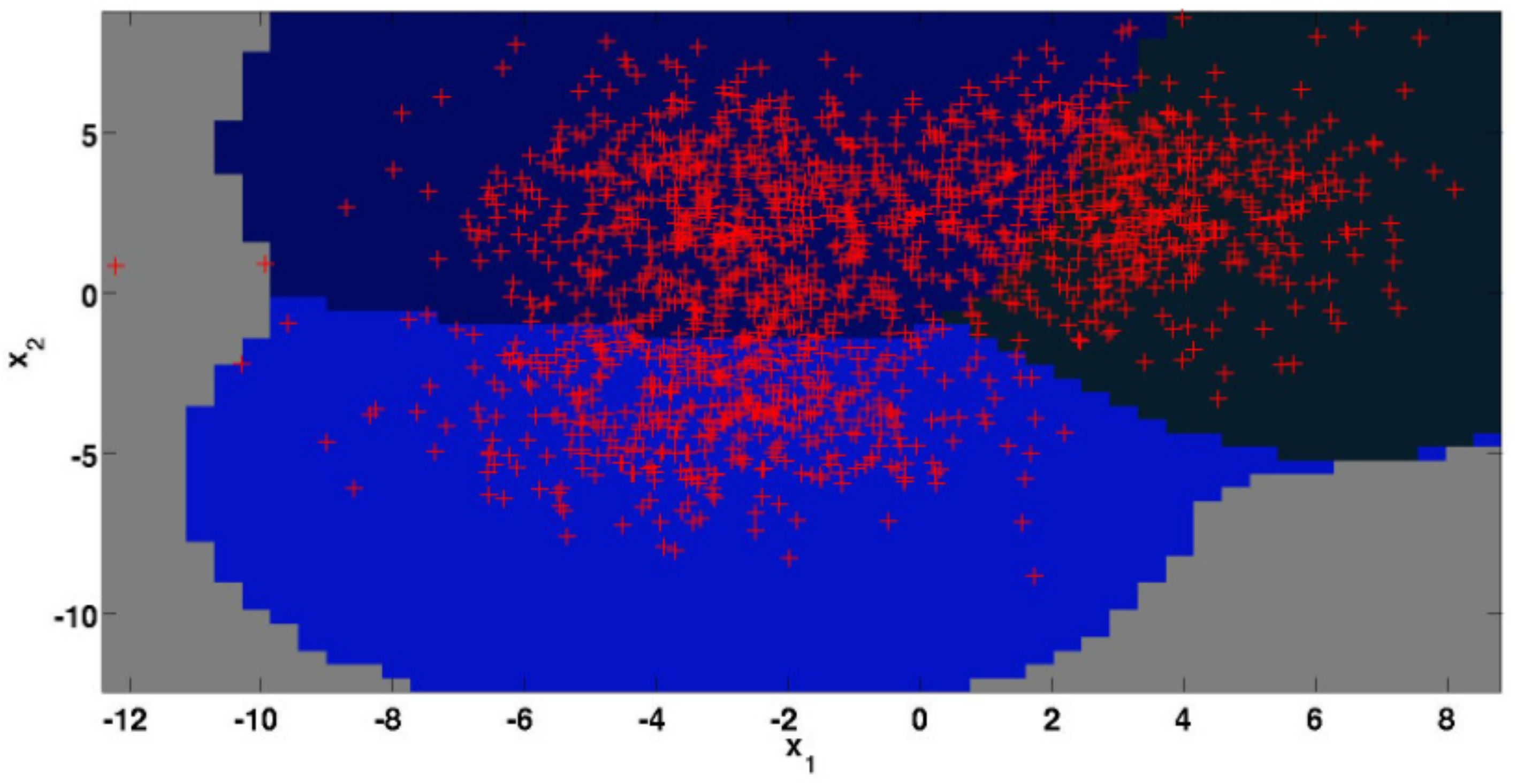}
\end{tabular}
\caption{\textbf{Sparse KSC on toy dataset}. \textbf{(Top)} Gaussian mixture with three highly overlapping components \textbf{(Center)} Clustering results, where the reduced set points are indicated with red circles \textbf{(Bottom)} Generalization boundaries.}\label{fig_glasso}
\end{figure}

In Figure \ref{fig_glasso} an example of clustering obtained using the group lasso formulation (\ref{grouplasso_sksc}) on a toy dataset is depicted. We can notice how the sparse KSC model is able to obtain high quality generalization using only $4$ points in the training set.

\section{Applications}
\label{app}
The KSC algorithm has been successfully used in a variety of applications in different domains. In the next Sections we will illustrate various results obtained in different fields such as computer vision, information retrieval and power load consumer segmentation.

\subsection{Image Segmentation}
\label{img_seg}
\begin{figure}[htbp]
\centering
\begin{tabular}{c}
\includegraphics[width=\textwidth]{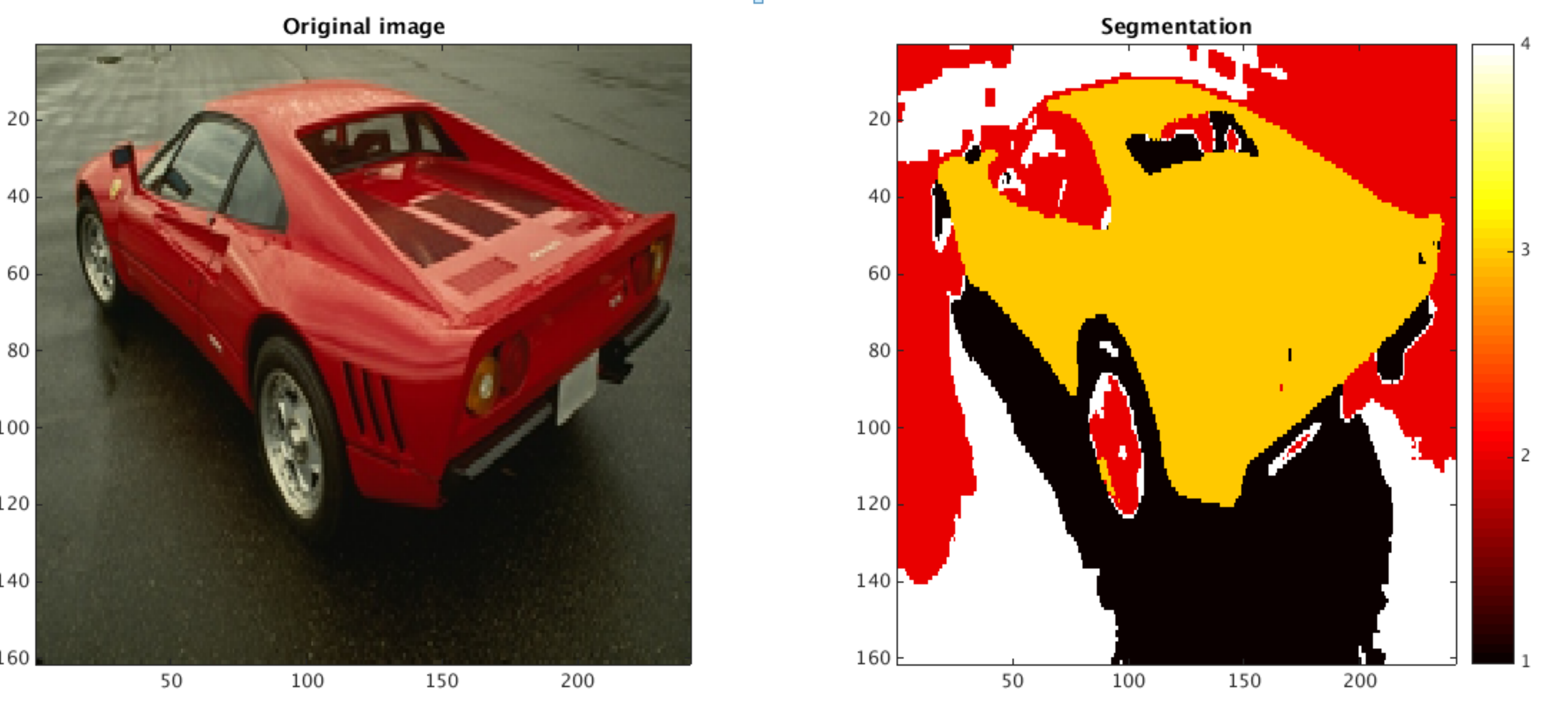}
\end{tabular}
\caption{\textbf{Image segmentation}. \textbf{(Left)} Original image \textbf{(Right)} Segmentation given by KSC.}\label{KSC_img}
\end{figure}

Image segmentation relates to partitioning a digital image into multiple regions, such that pixels in the same group share a certain visual content. In the experiments performed using KSC only the color information is exploited in order to segment the given images\footnote{The images have been extracted from the Berkeley image database \cite{berkeley_image}.}. More precisely, a local color histogram with a $5 \times 5$ pixels window around each pixel is computed using minimum variance color quantization of $8$ levels. Then, in order to compare the similarity between two histograms $h^{(i)}$ and $h^{(j)}$, the positive definite $\chi^2$ kernel $K(h^{(i)},h^{(j)}) = \exp(-\dfrac{\chi_{ij}^2}{\sigma_{\chi}^2})$ has been adopted \cite{sc_nystrom}. The symbol $\chi_{ij}^2$ denotes the $\chi_{ij}^2$ statistical test used to compare two probability distributions \cite{chi2test}, $\sigma_{\chi}$ as usual indicates the bandwidth of the kernel. In Figure \ref{KSC_img} an example of segmentation obtained using the basic KSC algorithm is given. 

\subsection{Scientific Journal Clustering}
\label{doc_clu}
We present here an integrated approach for clustering scientific journals using KSC. Textual information is combined with cross-citation information in order to obtain a coherent grouping of the scientific journals and to improve over existing journal categorizations.  The number of clusters $k$ in this scenario is fixed to $22$ since we want to compare the results with respect to the $22$ essential science indicators (ESI) shown in Table \ref{tab:clustering:22ESI}.
\begin{table}
	{\small
	\begin{center}
		\begin{tabular}{cc}
	\begin{tabular}{||c|c||}
		\hline
		Field & Name\\
		\hline
		\hline
		1 & Agricultural sciences \\
		2 & Biology and biochemistry\\
		3 & Chemistry\\
		4 & Clinical medicine\\
		5 & Computer science\\
		6 & Economics and business\\
		7 & Engineering\\
		8 & Environment/Ecology\\
		9 & Geosciences\\
		10 & Immunology\\
		11 & Materials sciences\\
		\hline
	\end{tabular}	&
	\begin{tabular}{||c|c||}
		\hline
		Field & Name\\
		\hline
		\hline
		12 & Mathematics\\
		13 & Microbiology\\
		14 & Molecular biology \& genetics\\
		15 & Multidisciplinary\\
		16 & Neuroscience \& behavior\\
		17 & Pharmacology \& toxicology\\
		18 & Physics\\
		19 & Plant \& animal science\\
		20 & Psychology / Psychiatry\\
		21 & Social sciences\\
		22 & Space science\\
		\hline
	\end{tabular}
	\end{tabular}
	\end{center}
	\caption{The $22$ science fields according to the essential science indicators (ESI)}
	\label{tab:clustering:22ESI}
	}
\end{table}

The data correspond to more than six million scientific papers indexed by the Web of Science (WoS) in the period $2002-2006$.  The type of manuscripts considered is article, letter, note and review. Textual information has been extracted from titles, abstracts and keywords of each paper together with citation information. From these data, the resulting number of journals under consideration is $8,305$.

The two resulting datasets contain textual and cross-citation information and are described as follows:
\begin{itemize}
	\item \textbf{Term/Concept by Journal dataset:}  The textual information was processed using the term frequency - inverse document frequency (TF-IDF) weighting procedure \cite{tfidf}.  Terms which occur only in one document and stop words were not considered into the analysis.   The Porter stemmer was applied to the remaining terms in the abstract, title and keyword fields.  This processing leads to a term-by-document matrix of around six million papers and $669,860$ term dimensionality.  The final journal-by-term dataset is a $8,305\times 669,860$ matrix. Additionally, latent semantic indexing (LSI) \cite{lsi} was performed on this dataset to reduce the term dimensionality to $200$ factors. 
	\item \textbf{Journal cross-citation dataset:} A different form of analyzing cluster information at the journal level is through a cross-citation graph.  This graph contains aggregated citations between papers forming a journal-by-journal cross-citation matrix.  The direction of the citations is not taken into account which leads to an undirected graph and a symmetric cross-citation matrix. 
\end{itemize}
The cross-citation and the text/concept datasets are integrated at the kernel level by considering the following linear combination of kernel matrices\footnote{Here we use the cosine kernel described in Table \ref{table_kernels}.}:
$$
\Omega^{\textrm{integr}}=\rho\Omega^{\textrm{cross-cit}}+(1-\rho)\Omega^{\textrm{text}}
$$
where $0\le\rho\le 1$ is a user-defined integration weight which value can be obtained from internal validation measures for cluster distortion\footnote{In our experiments we used the mean silhouette value (MSV) as an internal cluster validation criterion to select the value of $\rho$ which gives more coherent clusters.}, $\Omega^{\textrm{cross-cit}}$ is the cross-citation kernel matrix with $ij$-th entry  $\Omega^{\textrm{cross-cit}}_{ij}=K(x_i^{\text{cross-cit}},x_j^{\text{cross-cit}})$, $x_i^{\text{cross-cit}}$ is the $i$-th journal represented in terms of cross-citation variables, $\Omega^{\textrm{text}}$ is the textual kernel matrix with $ij$-th entry  $\Omega^{\textrm{text}}_{ij}=K(x_i^{\text{text}},x_j^{\text{text}})$, $x_i^{\text{text}}$ is the $i$-th journal represented in terms of textual variables and $i,j=1,\hdots,N$.

The KSC outcomes are depicted in Tables \ref{tab:clustering:journalresults} and \ref{tab:clustering:bestterms}. In particular, Table \ref{tab:clustering:journalresults} shows the results in terms of internal validation of cluster quality, namely mean silhouette value (MSV) \cite{silhouette} and Modularity \cite{newmanmod1,newmanmod2}, and in terms of agreement with existing categorizations (adjusted rand index or ARI \cite{arindex} and normalized mutual information (NMI \cite{nmijmlr}). Finally, Table \ref{tab:clustering:bestterms} shows the top $20$ terms per cluster, which indicate a coherent structure and illustrate that KSC is able to detect the text categories present in the corpus.  

\begin{table}
{\scriptsize
\begin{center}
  \begin{tabular}{c|c|c|c|c||c|c|c||}
	\cline{2-8}
	& \multicolumn{4}{||c||}{Internal validation} & \multicolumn{3}{|c||}{External validation}\\
	\cline{2-8}
	 & \multicolumn{1}{||c|}{MSV} & {MSV} & {MSV} & {Modularity} & Modularity & {ARI} & {NMI}\\
	& \multicolumn{1}{||c|}{textual} & cross-cit. & integrated & cross-cit. & ISI 254 & 22 ESI & 22 ESI\\
    \cline{2-8}
    \hline
	\cline{2-8}
    \multicolumn{1}{||c|}{$22$ ESI fields} & $0.057$ & $0.016$ & $0.063$ &$0.475$ & $0.526^\star$ & $1.000$ & $1.000$\\
    \hline
     \multicolumn{1}{||c|}{Cross-citations} & $0.093$ & $0.057$ & $0.189$ &$\mathbf{0.547}$ & $0.442$ & $0.278$ & $0.516$\\
    \hline
    \multicolumn{1}{||c|}{Textual (LSI)} & $0.118$ & $0.035$ & $0.130$ &$0.505$& $0.451$ &  $0.273$ & $0.516$\\
	\hline
    \multicolumn{1}{||c|}{Hierarch. Ward's method ${\rho}=0.5$} & $0.121$ & $0.055$ & $0.190$ &$\mathbf{0.547}$ & $\mathbf{0.488}$ & $0.285$ & $0.540$\\
     \hline
	 \hline
    \multicolumn{1}{||c|}{Integr. Terms+Cross-citations ${\rho}=0.5$} & $0.138$ &	$\mathbf{0.064}$	 & $\mathbf{0.201}$ &$\mathit{0.533}$	&	$\mathit{0.465}$	& $0.294$ &	$\mathit{0.557}$ \\
     \hline
    \multicolumn{1}{||c|}{Integr. LSI+Cross-citations ${\rho}=0.5$} & $\mathit{0.145}$ & $\mathit{0.062}$ & $\mathit{0.197}$ &$0.527$ & $\mathit{0.465}$  & $0.308$ & $\mathbf{0.560}$\\
    \hline
  \end{tabular}
\end{center}
}
\caption{\textbf{Text clustering quality.} Spectral clustering results of several integration methods in terms of mean Silhouette value (MSV), modularity, adjusted Rand index (ARI) and normalized mutual information (NMI). The first four rows correspond to existing clustering results used for comparison.  The last two rows correspond to the proposed spectral clustering algorithms.  For external validation, the clustering results are compared with respect to the 22 ESI fields and the ISI 254 subject categories. The highest value per column is indicated in bold while the second highest value appears in italic. For MSV, a standard t-test for the difference in means revealed that differences between highest and second highest values are statistically significant at the $1\%$ significance level ($p$-value $< 10^{–8}$).  The selected method for further comparisons is the integrated LSI+Cross-citations approach since it wins in external validation with one highest value (NMI) and one second highest value (Modularity).}
\label{tab:clustering:journalresults}
\end{table}

\begin{table}
{\scriptsize
\begin{center}
\begin{tabular}{cc}
\begin{tabular}{p{10mm} p{44mm}}
\toprule
&\multicolumn{1}{c}{{Best $20$ terms}} \\
\hline
\multirow{4}{*}{Cluster $1$} & diabet therapi hospit arteri coronari physician renal hypertens mortal syndrom cardiac nurs chronic infect pain cardiovascular symptom serum cancer pulmonari\\
\hline
\multirow{4}{*}{Cluster $2$} & polit war court reform parti legal gender urban democraci democrat civil capit feder discours economi justic privat liber union welfar\\
\hline
\multirow{3}{*}{Cluster $3$} & diet milk fat intak cow dietari fed meat nutrit fatti chees vitamin ferment fish dry fruit antioxid breed pig egg \\
\hline
\multirow{4}{*}{Cluster $4$} & alloi steel crack coat corros fiber concret microstructur thermal weld film deform ceram fatigu shear powder specimen grain fractur glass\\
\hline
\multirow{4}{*}{Cluster $5$} & infect hiv vaccin viru immun dog antibodi antigen pathogen il pcr parasit viral bacteri dna therapi mice bacteria cat assai\\
\hline
\multirow{4}{*}{Cluster $6$} & psycholog cognit mental adolesc emot symptom child anxieti student sexual interview school abus psychiatr gender attitud mother alcohol item disabl\\
\hline
\multirow{4}{*}{Cluster $7$} & text music polit literari philosophi narr english moral book essai write discours philosoph fiction ethic poetri linguist german christian religi\\
\hline
\multirow{4}{*}{Cluster $8$} & firm price busi trade economi invest capit tax wage financi compani incom custom sector bank organiz corpor stock employ strateg\\
\hline
\multirow{4}{*}{Cluster $9$} & nonlinear finit asymptot veloc motion stochast elast nois turbul ltd vibrat iter crack vehicl infin singular shear polynomi mesh fuzzi\\
\hline
\multirow{3}{*}{Cluster $10$} & soil seed forest crop leaf cultivar seedl ha shoot fruit wheat fertil veget germin rice flower season irrig dry weed\\
\hline
\multirow{4}{*}{Cluster $11$} & soil sediment river sea climat land lake pollut wast fuel wind ocean atmospher ic emiss reactor season forest urban basin\\
\bottomrule
\end{tabular} &
\begin{tabular}{p{10mm} p{46mm}}
\toprule
&\multicolumn{1}{c}{Best $20$ terms} \\
\hline
\multirow{4}{*}{Cluster $12$} & algebra theorem manifold let finit infin polynomi invari omega singular inequ compact lambda graph conjectur convex proof asymptot bar phi\\
\hline
\multirow{3}{*}{Cluster $13$} & pain surgeri injuri lesion muscl bone brain ey surgic nerv mri ct syndrom fractur motor implant arteri knee spinal stroke\\
\hline
\multirow{4}{*}{Cluster $14$} & rock basin fault sediment miner ma tecton isotop mantl volcan metamorph seismic sea magma faci earthquak ocean cretac crust sedimentari\\
\hline
\multirow{4}{*}{Cluster $15$} & web graph fuzzi logic queri schedul semant robot machin video wireless neural node internet traffic processor retriev execut fault packet\\
\hline
\multirow{4}{*}{Cluster $16$} & student school teacher teach classroom instruct skill academ curriculum literaci learner colleg write profession disabl faculti english cognit peer gender\\
\hline
\multirow{3}{*}{Cluster $17$} & habitat genu fish sp forest predat egg nest larva reproduct taxa bird season prei nov ecolog island breed mate genera\\
\hline
\multirow{4}{*}{Cluster $18$} & star galaxi solar quantum neutrino orbit quark gravit cosmolog decai nucleon emiss radio nuclei relativist neutron cosmic gaug telescop hole\\
\hline
\multirow{4}{*}{Cluster $19$} & film laser crystal quantum atom ion beam si nm dope thermal spin silicon glass scatter dielectr voltag excit diffract spectra\\
\hline
\multirow{4}{*}{Cluster $20$} &
polym catalyst ion bond crystal solvent ligand hydrogen nmr molecul atom polymer poli aqueou adsorpt methyl film spectroscopi electrod bi\\
\hline
\multirow{4}{*}{Cluster $21$}&
receptor rat dna neuron mice enzym genom transcript brain mutat peptid kinas inhibitor metabol cancer mrna muscl ca2 vitro chromosom\\
\hline
\multirow{4}{*}{Cluster $22$} & cancer tumor carcinoma breast therapi prostat malign chemotherapi tumour surgeri lesion lymphoma pancreat recurr resect surgic liver lung gastric node\\
\bottomrule
\end{tabular}
\end{tabular}
\end{center}
}
\caption{\textbf{Text clustering results.} Best $20$ terms per cluster according to the integrated results (LSI+cross-citation) with $\rho=0.5$.  The terms found display a coherent structure in the clusters.}
\label{tab:clustering:bestterms}
\end{table}

\subsection{Power Load Clustering}
\label{pl_clu}
\begin{figure}[htbp]
\centering
\begin{tabular}{c}
\includegraphics[width=\textwidth]{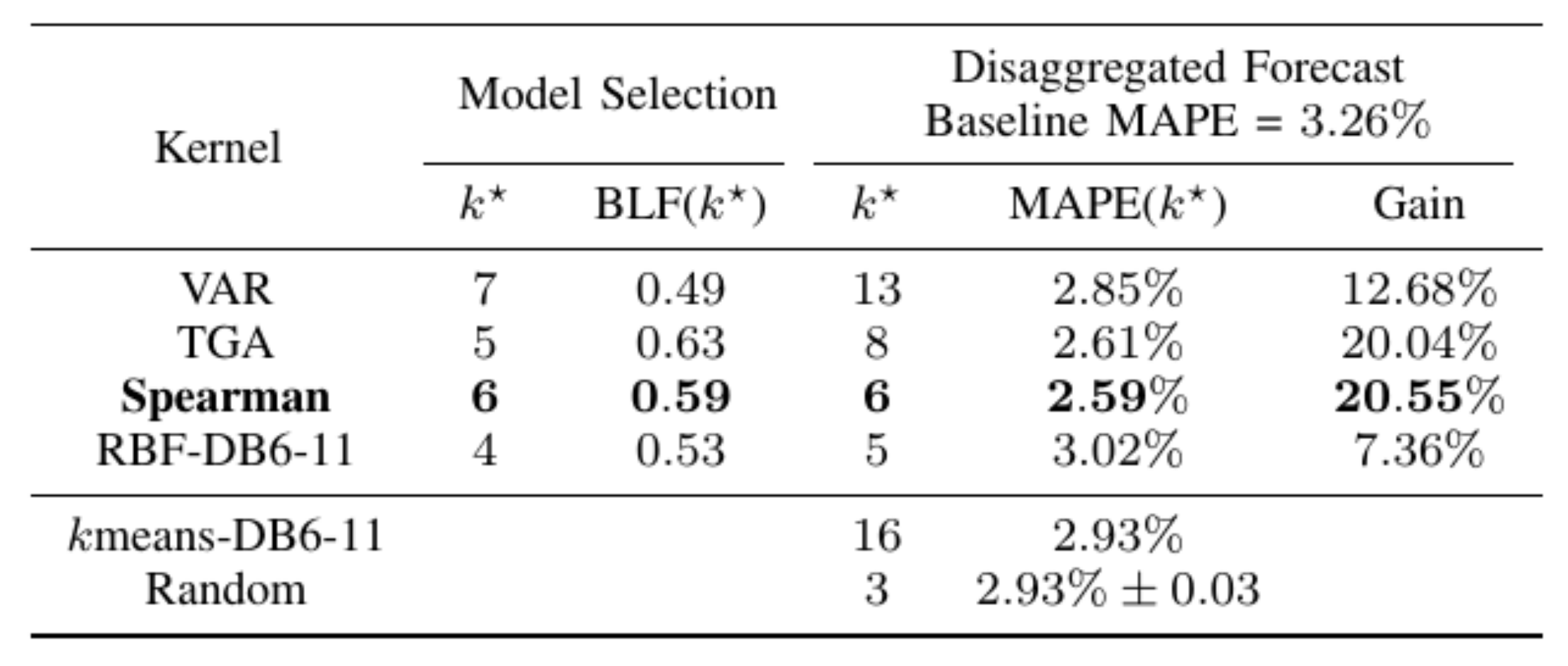}
\end{tabular}
\caption{\textbf{Kernel comparisons for power load clustering data}. Model selection and forecasting results in terms of the mean absolute percentage error (MAPE).  RBF-DB6-11 refers to using the RBF kernel on the detail coefficients using wavelets (DB6, 11 levels).  The winner is the Spearman-based kernel with a improvement of $20.55\%$.  For this kernel, the number of clusters $k$ found by the BLF also coincides with the number of aggregators needed to maximize the improvement. }  
\label{pl_tab}
\end{figure}

Accurate power load forecasts are essential in electrical grids and markets particularly for planning and control operations \cite{alzate_icann2009}. In this scenario, we apply KSC for finding power load smart meter data that are similar in order to aggregate them and improve the forecasting accuracy of the global consumption signal.  The idea is to fit a forecasting model on the aggregated load of each cluster (aggregator).  The $k$ predictions are summed to form the final disaggregated prediction.  The number of clusters and the time series used for each aggregator are determined via KSC \cite{Carlos_ICDM2013}.  The forecasting model used is a periodic autoregresive model with exogenous variables (PARX) \cite{galloclustering}.  Table \ref{pl_tab} (taken from \cite{Carlos_ICDM2013} shows the model selection and disaggregation results.  Several kernels appropriate for time series were tried including a Vector Autoregressive (VAR) kernel [Add: Cuturi, Autoregressive kernels for time series, arXiv], Triangular Global Alignment (TGA) kernel [Add: Cuturi, Fast Global Alignment Kernels, ICML 2011] and an RBF kernel with Spearman's distance.  The results show an improvement of $20.55\%$ with the similarity based on Spearman's corrleation in the forecasting accuracy compared to not using clustering at all (i.e., aggregating all smart meters).  The BLF was also able to detect the number of clusters that maximize the improvement (6 clusters in this case).  

\begin{figure}[htbp]
\centering
\begin{tabular}{c}
\includegraphics[width=\textwidth]{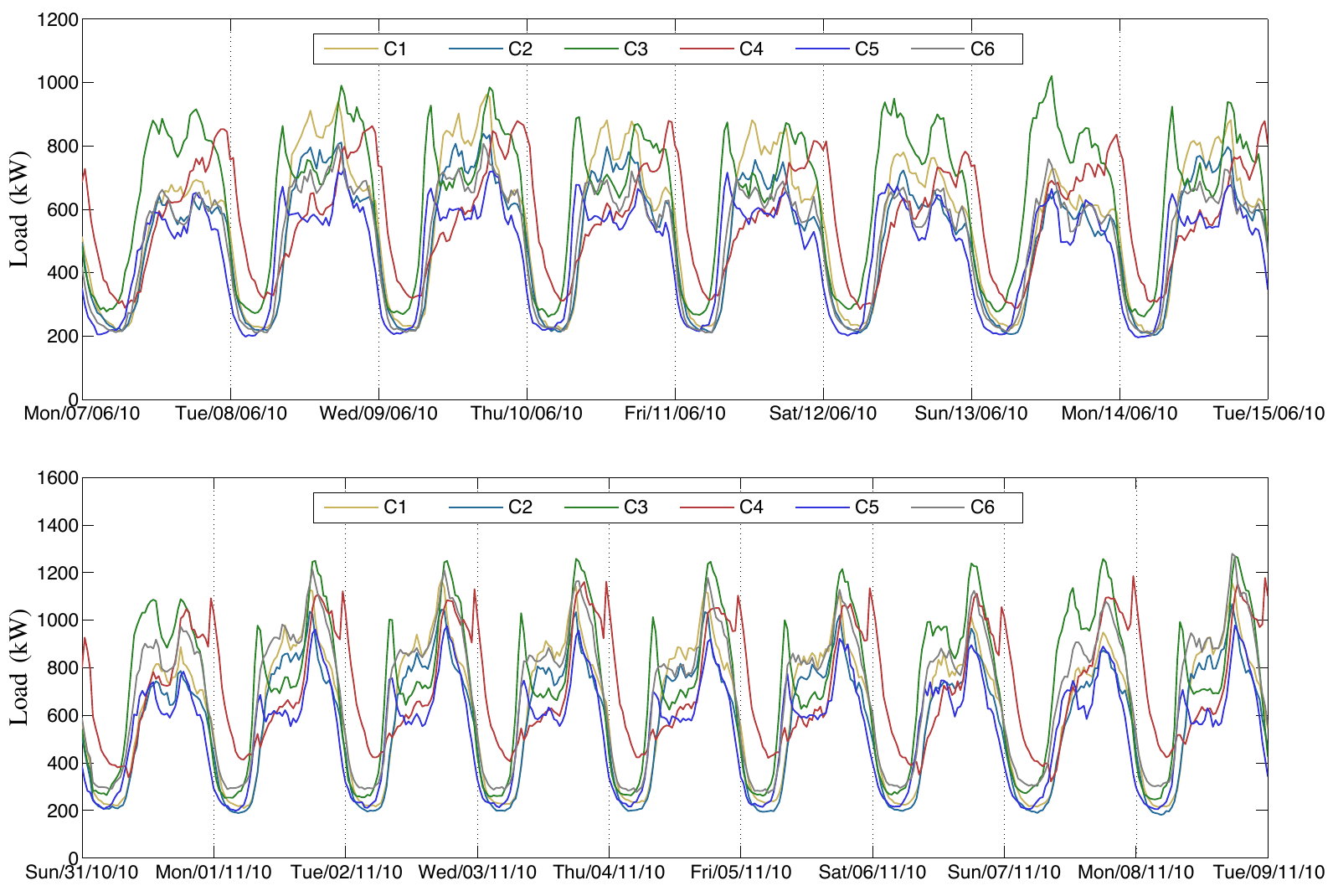}
\end{tabular}
\caption{\textbf{Power load clustering results}. Visualization of the $6$ clusters obtained by KSC. \textbf{(Top)} Aggregated load in summer. \textbf{(Bottom)} Aggregated load in winter. The daily cycles are clearly visible and the clusters capture different characteristics of the consumption pattern.  This clustering result improves the forecasting accuracy by $20.55\%$}  
\label{pl_img}
\end{figure}

\subsection{Big data}
\label{bigdata}
KSC has been shown to be effective in handling big data at a desktop PC scale. In particular, in \cite{raghvendra_bigdata}, we focused on community detection in big networks containing millions of nodes and
several million edges, and we explained how to scale our method by means of three steps\footnote{A \textit{Matlab} implementation of the algorithm can be downloaded at:\\ \textit{http://www.esat.kuleuven.be/stadius/ADB/mall/softwareKSCnet.php}}. First, we select a smaller subgraph that preserves the overall community structure by using the FURS algorithm \cite{FURS_paper}, where hubs in dense regions of the original graph are selected via a greedy activation-deactivation procedure. In this way the kernel matrix related to subgraph fits the main memory and the KSC model can be quickly trained by solving a smaller eigenvalue problem. Then the BAF criterion described in Section \ref{modsel_KSC}, which is memory and computationally efficient, is used for model selection\footnote{In \cite{raghvendra_sclara} this model selection step has been eliminated by proposing a self tuned method where the structure of the projections in the eigenspace is exploited to automatically identify an optimal cluster structure.}. Finally, the out-of-sample extension is used to infer the cluster memberships for the remaining nodes forming the test set (which is divided into chunks due to memory constraints).

\begin{figure}[!htbp]
\centering
\begin{tabular}{c}
\includegraphics[width=\textwidth]{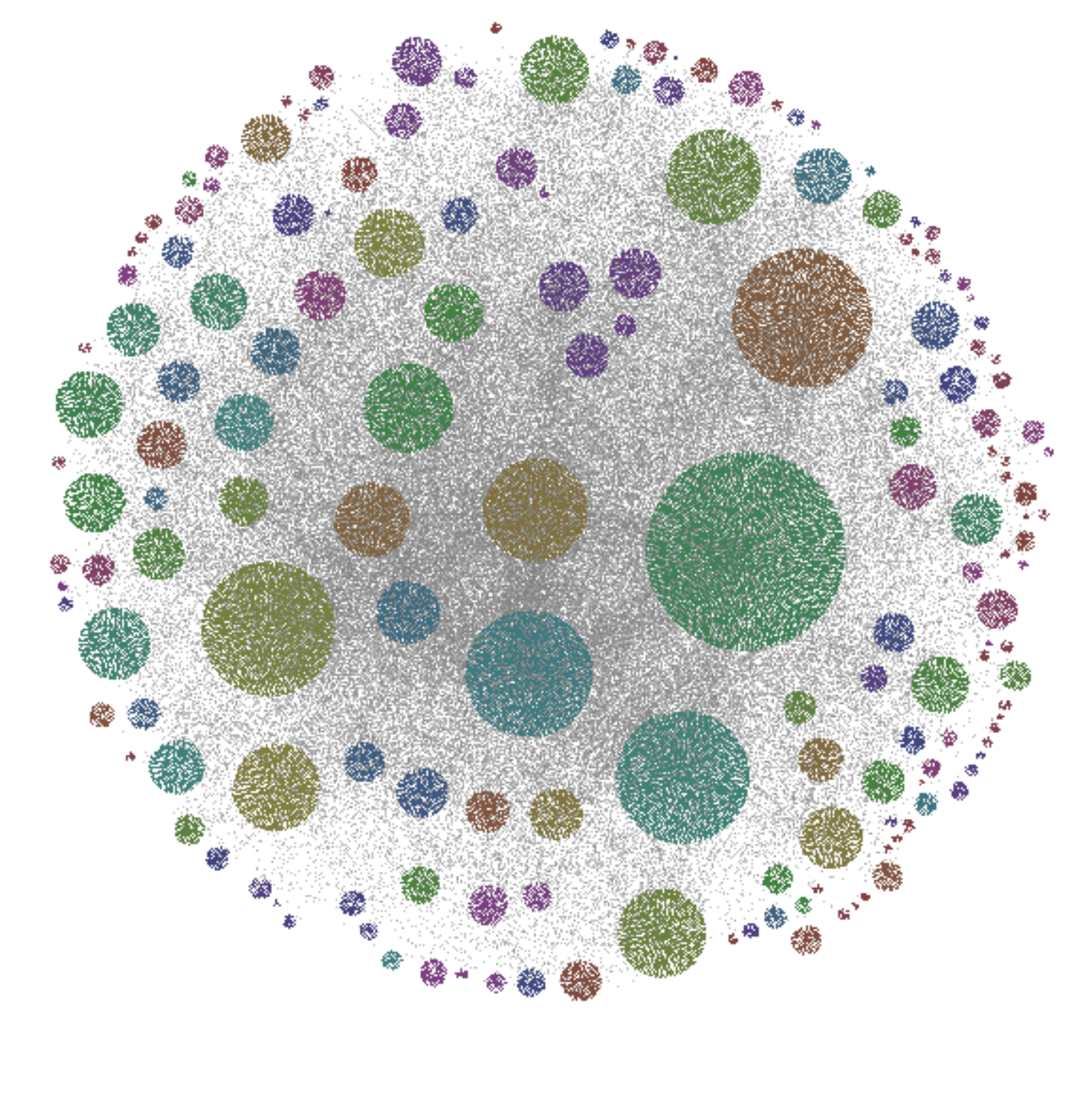}
\end{tabular}
\caption{\textbf{Large scale community detection}. Community structure detected at one particular hierarchical level by the AH-KSC method summarized in Section \ref{HC2}, related to the \textit{Cond-Mat} collaboration network.}  
\label{big_dat}
\end{figure}

In \cite{raghvendra_plosone} the hierarchical clustering technique summarized in Section \ref{HC2} has been used to perform community detection in real-life networks at different resolutions. The method has been shown to be able to detect complex structures at various hierarchical levels, by not suffering of any resolution limit. An example of results obtained on the  \textit{Cond-mat} network of collaborations between authors of papers submitted to Condense Matter category in \textit{Arxiv} \cite{Leskovec:2007} is shown in Figure \ref{big_dat}.

\begin{figure}[!htbp]
\centering
\begin{tabular}{c}
\includegraphics[width=\textwidth]{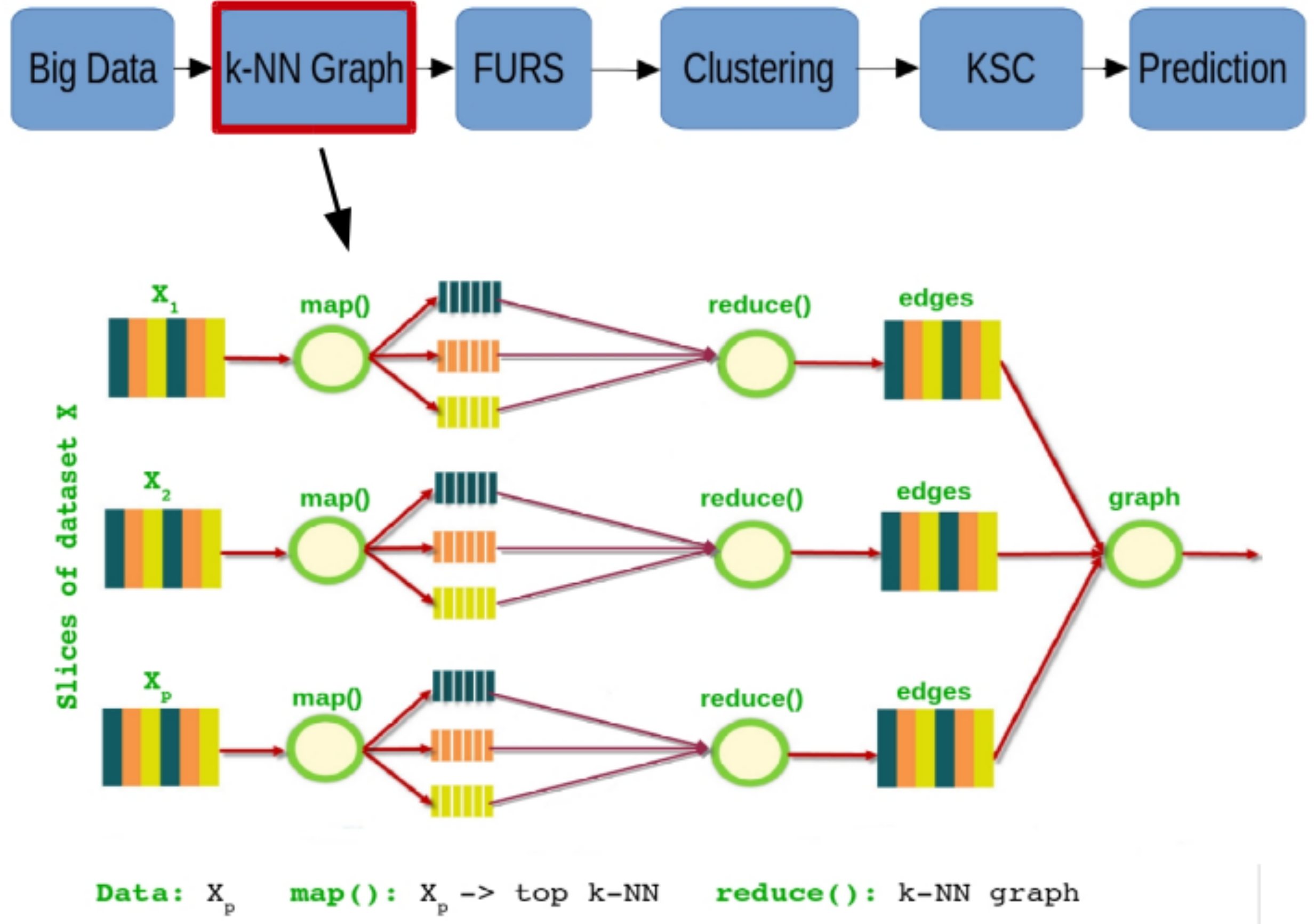}
\end{tabular}
\caption{\textbf{Big data clustering}. \textbf{(Top)} Illustration of the steps involved in clustering big vector data using KSC. \textbf{(Bottom)} Map-Reduce procedure used to obtain a representative training subset by constructing a k-NN graph.}  
\label{map_red}
\end{figure}

Finally, in \cite{raghvendra_bigdata2014}, we propose a deterministic method to obtain subsets from big vector data which are a good representative of the inherent clustering structure. We first convert the large scale dataset into a sparse undirected k-NN graph using a Map-Reduce framework. Then, the FURS method is used to select a few representative nodes from this graph, corresponding to certain data points in the original dataset. These points are then used to quickly train the KSC model, while the generalization property of the method is exploited to compute the cluster memberships for the remainder of the dataset. In Figure \ref{map_red} a summary of all these steps is sketched.

\section{Conclusions}
\label{conclusion_ch_langonetal}
In this chapter we have discussed the kernel spectral clustering (KSC) method, which is cast in an LS-SVM learning framework. We have explained that, like in the classifier case, the clustering model can be trained on a subset of the data with optimal tuning parameters, found during the validation stage. The model is then able to generalize to unseen test data thanks to its out-of-sample extension property. Beyond the core algorithm, some extensions of KSC allowing to produce probabilistic and hierarchical outputs have been illustrated. Furthermore, two different approaches to sparsify the model based on the Incomplete Cholesky Decomposition (ICD) and $L_1$ and $L_0$ penalties have been described. This allows to handle large scale data at a desktop scale. Finally, a number of applications in various fields ranging from computer vision to text mining have been examined.

\begin{acknowledgement}
EU: The research leading to these results has received funding from the European Research Council under the European Union's Seventh Framework Programme (FP7/2007-2013) / ERC AdG A-DATADRIVE-B (290923). This chapter reflects only the authors' views, the Union is not liable for any use that may be made of the contained information. Research Council KUL: GOA/10/09 MaNet, CoE PFV/10/002 (OPTEC), BIL12/11T; PhD/Postdoc grants. Flemish Government: FWO: projects: G.0377.12 (Structured systems), G.088114N (Tensor based data similarity); PhD/Postdoc grants. IWT: projects: SBO POM (100031); PhD/Postdoc grants. iMinds Medical Information Technologies SBO 2014. Belgian Federal Science Policy Office: IUAP P7/19 (DYSCO, Dynamical systems, control and optimization, 2012-2017.)
\end{acknowledgement}

%
%
     
\bibliographystyle{jphysicsB}
\bibliography{bib_phdthesis}

\include{appendix}

\backmatter
\include{glossary}
\include{solutions}
\printindex


\end{document}